\documentclass[11pt]{article}
\usepackage[a4paper, bottom=1in, top=1.2in, left=1in, right=1in]{geometry}

\usepackage{times}

\usepackage{amsthm}
\usepackage{amssymb}
\usepackage{enumitem} 
\usepackage{amsmath}

\usepackage[bookmarks=true]{hyperref}
\hypersetup{%
    colorlinks=true,       
    linkcolor=blue,       
    citecolor=blue,       
    filecolor=black,        
    urlcolor=purple,        
    linktoc=page            
}

\usepackage{tikz}
\usetikzlibrary{arrows.meta}
\usetikzlibrary{
    fit,shapes,
    shapes.geometric,
    shapes.symbols,
    shapes.arrows,
    shapes.multipart,
    shapes.callouts,
    shapes.misc}

\usepackage{orcidlink}

\usepackage{tcolorbox}

\usepackage{breqn}
\usepackage{booktabs}

\usepackage[labelfont=bf]{caption}

\usepackage{wrapfig}
\usepackage{graphicx}
\graphicspath{ {images/} }
\usepackage{subcaption}

\usepackage{array}
\newcolumntype{P}[1]{>{\centering\arraybackslash}p{#1}} 
\newcolumntype{M}[1]{>{\centering\arraybackslash}m{#1}} 
\usepackage{multirow}

\makeindex

\title{ASP-Assisted Symbolic Regression:\\ Uncovering Hidden Physics in Fluid Mechanics}

\author{
Theofanis Aravanis\textsuperscript{1, 2, \orcidlink{0000-0003-0329-3200}} \and 
Grigorios Chrimatopoulos\textsuperscript{2, \orcidlink{0009-0006-7413-6316}} \and
Mohammad Ferdows\textsuperscript{3, \orcidlink{0000-0002-8696-9116}} \and 
Michalis Xenos\textsuperscript{4, \orcidlink{0000-0001-8441-1306}} \and
Efstratios Em. Tzirtzilakis\textsuperscript{5, \orcidlink{0000-0002-5598-564X}}
}

\date{\small
\textsuperscript{1}Department of Digital Systems, University of the Peloponnese, Sparta 231 00, Greece\\[0.6ex]
\textsuperscript{2}Department of Mechanical Engineering, University of the Peloponnese, Patras 263 34, Greece\\[0.6ex]
\textsuperscript{3}Department of Applied Mathematics, University of Dhaka, Dhaka-1000, Bangladesh\\[0.6ex]
\textsuperscript{4}Department of Mathematics, University of Ioannina, Ioannina 451 10, Greece\\[0.6ex]
\textsuperscript{5}Department of Civil Engineering, University of the Peloponnese, Patras 263 34, Greece\\[1.5ex]
$\dagger\, $e-mails: 
T. Aravanis (\href{mailto:taravanis@go.uop.gr}{taravanis@go.uop.gr}), 
G. Chrimatopoulos (\href{mailto:g.chrimatopoulos@go.uop.gr}{g.chrimatopoulos@go.uop.gr}),\\ 
M. Ferdows (\href{mailto:ferdows@du.ac.bd}{ferdows@du.ac.bd}), 
M. Xenos (\href{mailto:mxenos@uoi.gr}{mxenos@uoi.gr}), E. E. Tzirtzilakis (\href{mailto:etzirtzilakis@go.uop.gr}{etzirtzilakis@go.uop.gr})
}

\usepackage{fancyhdr}
\providecommand{\keywords}[1]{\textbf{\textit{Keywords:}} #1}

\begin{document}

\maketitle
\sloppy

\begin{abstract}
Symbolic Regression (SR) offers an interpretable alternative to conventional Machine-Learning (ML) approaches, which are often criticized as ``black boxes''. In contrast to standard regression models that require a prescribed functional form, SR constructs expressions from a user-defined set of mathematical primitives, enabling the automated discovery of compact formulas that fit the data and reveal underlying physical relationships. In fluid mechanics, where understanding the underlying physics is as crucial as predictive accuracy, this study applies SR to model three-dimensional (3D) laminar flow in a rectangular channel, focusing on the axial velocity and pressure fields. Compact symbolic equations were derived from numerical simulation data, accurately reproducing the expected parabolic velocity profile and linear pressure drop, and showing excellent agreement with analytical solutions from the literature. To address the limitation that purely data-driven SR models may overlook domain-specific constraints, an innovative hybrid framework that integrates SR with Answer Set Programming (ASP) is also introduced. This integration combines the generative power of SR with the declarative reasoning capabilities of ASP, ensuring that derived equations remain both statistically accurate and physically plausible. The proposed SR/ASP methodology demonstrates the potential of combining data-driven and knowledge-representation approaches to enhance interpretability, reliability, and alignment with physical principles in fluid dynamics and related domains.
\end{abstract}

\vspace{4mm}


\newpage

\section{Introduction}

Fluid mechanics is essential to many fields, from engineering and environmental science to medical research. However, the complexity of fluid behaviour, especially in turbulent or multiphase systems, makes it challenging to model and understand. Traditional methods, including Computational Fluid Dynamics (CFD), have been highly effective and physically transparent, yet often incur substantial computational cost. On the other hand, modern Machine-Learning (ML) approaches range from interpretable methods to more opaque models such as Artificial Neural Networks. Although ML surrogates are often used in fluid mechanics primarily for computational efficiency, many commonly adopted models still behave as ``black boxes'', offering limited physical interpretability --- an important limitation in applications where understanding the underlying flow physics is as important as prediction quality.

Against this background, research increasingly turns to data-driven methods for new solutions, and approaches that combine accuracy with clear, understandable models. {\em Symbolic Regression} (SR) stands out in this space. Unlike typical ML models that operate as ``black boxes'', SR identifies mathematical expressions that describe relationships in the data. Rather than relying on a fixed predefined functional form, SR searches over combinations of user-specified mathematical primitives, enabling it to reveal or confirm physically meaningful relationships in fluid mechanics \cite{kronberger24}. This feature is invaluable for researchers and engineers who need models that are not just predictive, but also provide insight into the underlying physics.

Despite the interpretability of SR and its successful use in a plethora of domains \cite{makke24,angelis23,schmidt09,udrescu20,kim21}, the application of evolutionary algorithm-based SR in fluid mechanics remains relatively limited. In what follows, we briefly review representative studies that apply SR within fluid-mechanics contexts.

Praks and Brki{\'c} \cite{praks18} demonstrated its potential by approximating the Colebrook equation, a key relation for calculating turbulent flow friction factors. By leveraging extensive datasets of Reynolds numbers and relative roughness, the initial approximations produced by SR tools were iteratively refined using fixed-point methods. This iterative process underscored the adaptability of SR in improving accuracy while maintaining low computational complexity.

Further advancements in SR applications include studies by El Hasadi and Padding \cite{hasadi19}, who utilized semi-supervised guided symbolic regression to explore the fluid drag experienced by ellipsoidal and spherocylindrical particles with arbitrary aspect ratios. Their work contributed novel correlations for drag forces, broadening the scope of SR in particle-fluid interactions. Milo{\v{s}}evi{\'c} {\em et al.} \cite{milosevic22} contributed to SR's growing relevance by combining it with data-driven insights to unify models of laminar and turbulent flow friction, proposing novel formulations that efficiently bridged these flow regimes. Sofos {\em et al.} \cite{sofos22} also employed SR to define a Lennard-Jones fluid descriptor, correlating density and temperature variables, further emphasizing SR's adaptability across diverse fluid systems. Roland {\em et al.} \cite{roland19} explored SR in non-Newtonian fluid mechanics, creating accurate models for viscous dissipation in polymer extrusion processes. Their work reduced reliance on computationally expensive numerical simulations, showcasing SR's role in optimizing industrial processes.

Reassessing transport properties of Lennard-Jones fluids, Angelis {\em et al.} \cite{angelis24} employed SR to develop closed-form equations for viscosity and thermal conductivity. These models encompassed transitions from dilute gases to dense liquids, achieving fine accuracy with reduced complexity compared to traditional molecular dynamics simulations. Similarly, Alam {\em et al.} \cite{alam22} used SR to predict self-diffusion in Lennard-Jones fluids, finding SR-derived models more interpretable and accurate than many empirical relationships.

In turbulence modelling, Wu and Zhang \cite{wu23} enhanced the predictive power of the shear-stress-transport (SST) turbulence model using SR to derive correction terms that generalize well across diverse test cases, including three-dimensional flows. This approach demonstrated SR's ability to overcome the generalization limitations of machine-learning-based turbulence models. Chakrabarty and Yakovenko \cite{chakrabarty21} applied SR to improve the Reynolds stress approximations within Reynolds-averaged Navier-Stokes (RANS) equations, emphasizing its potential to derive compact, interpretable models for turbulence closures.

In experimental setups, Reinbold {\em et al.} \cite{reinbold21} showcased SR's capacity to distil meaningful equations from high-dimensional, noisy data by incorporating prior physical knowledge. Their study reconstructed external force fields absent from the training data, illustrating how SR can leverage incomplete information to generate robust models. This capability demonstrates SR's strength in aligning empirical observations with theoretical frameworks, yielding models that closely approximate foundational equations like the Navier-Stokes equations.

\begin{figure}[h!]
\centering\small
\scalebox{1}{
\begin{tikzpicture}[very thick, node distance=3cm]
\node (input1) [draw, shape=rounded rectangle, align=center, minimum width=2.5cm, minimum height=1.5cm, text centered] at (-4, 0) {Input\\ Data};
\node (method) [draw, shape=rectangle, rounded corners, align=center, minimum width=2.5cm, minimum height=1.5cm, text centered] at (0, 0) {Numerical\\ Method};
\node (output1) [draw, shape=rounded rectangle, align=center, minimum width=2.5cm, minimum height=1.5cm, text centered] at (5, 0) {Numerical\\ Solution};

\node (input2) [draw, shape=rounded rectangle, align=center, minimum width=2.5cm, minimum height=1.5cm, text centered] at (-4, -2.8) {Input\\ Data};
\node (sr) [draw, shape=rectangle, rounded corners, align=center, minimum width=2.5cm, minimum height=1.5cm, text centered] at (0,-2.8) {Symbolic \\ Expressions};
\node (output2) [draw, shape=rounded rectangle, align=center, minimum width=4.5cm, minimum height=2cm, text centered] at (5, -2.8) {Rapid, Accurate\\ and Interpretable \\ Prediction of the\\ Numerical Solution};

\draw[->, >=stealth, line width=0.5mm] (input1) -- (method);
\draw[->, >=stealth, line width=0.5mm] (method) -- (output1);
\draw[->, >=stealth, line width=0.5mm] (input2) -- (sr);
\draw[->, >=stealth, line width=0.5mm] (sr) -- (output2);
\end{tikzpicture}}
\caption{The SR approach (bottom) can operate, through its derived symbolic expressions, as an efficient, interpretable surrogate for traditional, computationally intensive numerical methods (top).}
\label{fig_sr_approach}
\end{figure}
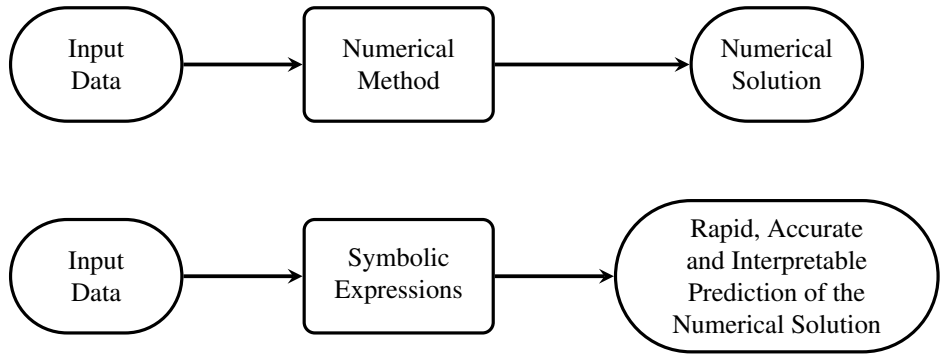

In this article, we push the frontier of fluid-dynamics modelling by harnessing SR to extract interpretable, closed-form expressions that capture the essential physics of a fundamental fluid-flow problem. Our approach establishes a novel and robust benchmark for interpretable ML models in viscous fluid motion within confined geometries. Instead of prescribing a specific functional form, our SR framework successfully derives explicit expressions for the (axial) \emph{velocity} and \emph{pressure} fields in \emph{three-dimensional} (3D) channels as functions of {\em spatial coordinates} and the {\em Reynolds number}. As demonstrated, the derived symbolic expressions closely match the corresponding numerical solutions, reproducing the expected physical behaviour and aligning with established analytical trends. These results suggest that the SR-based expressions offer an interpretable and computationally efficient representation of the underlying flow physics, while remaining consistent with high-fidelity numerical predictions (see 
Figure~\ref{fig_sr_approach}).

\vspace{3mm}

\begin{figure}[t]
\centering\small
\scalebox{1}{
\begin{tikzpicture}[very thick, node distance=3cm]
\small
\node (data) [draw, shape=rounded rectangle, align=center, minimum width=3.2cm, minimum height=1.5cm, text centered] at (-0.2,0) {Fluid-Mechanics\\ Raw Data};
\node (sr) [draw, shape=rectangle, rounded corners, align=center, minimum width=2.3cm, minimum height=1.5cm, text centered] at (3,0) {SR Module};
\node (sr_out) [draw, shape=rounded rectangle, align=center, minimum width=3cm, minimum height=2cm, text centered] at (6,0) {Accurate\\ Symbolic\\ Expressions};
\node (asp) [draw, shape=rectangle, rounded corners, align=center, minimum width=2.3cm, minimum height=1.5cm, text centered] at (9,0) {ASP Module};
\node (asp_out) [draw, shape=rounded rectangle, align=center, minimum width=3.5cm, minimum height=2cm, text centered] at (12.3,0) {Accurate \&\\ Physically Valid\\ Symbolic\\ Expressions};
\node (domain) [draw, shape=tape, align=center, minimum width=2.6cm, minimum height = 1.5cm] at (9,2.5) {Domain-Specific\\ Constraints};

\draw[->, >=stealth, line width=0.5mm] (data) -- (sr);
\draw[->, >=stealth, line width=0.5mm] (sr) -- (sr_out);
\draw[->, >=stealth, line width=0.5mm] (sr_out) -- (asp);
\draw[->, >=stealth, line width=0.5mm] (asp) -- (asp_out);
\draw[->, >=stealth, line width=0.5mm] (domain) -- (asp);
\end{tikzpicture}}
\caption{Integrated SR and ASP work-flow. SR generates candidate accurate symbolic expressions from fluid-mechanics raw data. Then, ASP applies domain-specific constraints to filter and select expressions that are both accurate and physically valid.}
\label{fig_sr_asp}
\end{figure}
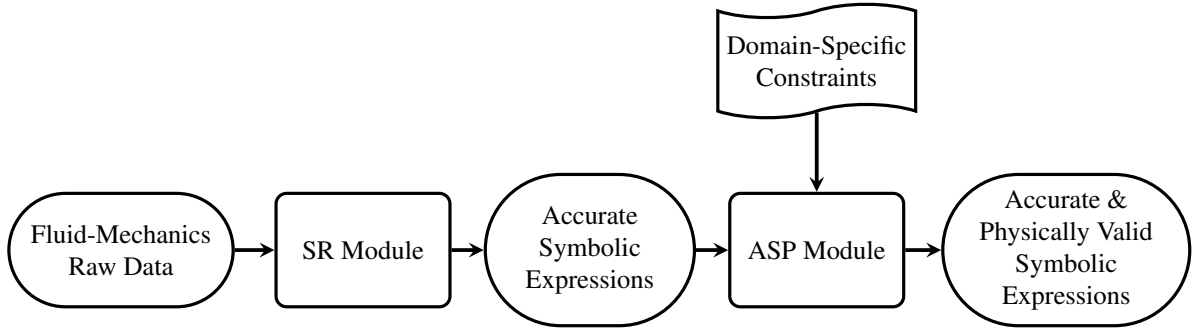

While, as we show, SR excels in its generative capabilities, it operates within a {\em purely data-driven} manner, potentially overlooking intricate domain-specific constraints and logical relationships inherent to physical systems. This limitation can lead to the selection of models that, despite their statistical accuracy, {\em may violate} fundamental physical laws or, in general, domain-specific constraints. To address these challenges, we propose an exploratory integration of SR with the {\em knowledge-representation} framework of {\em Answer Set Programming} (ASP) \cite{lifschitz19,brewka11,gebser12}. ASP constitutes a declarative-programming paradigm that allows the specification of complex problems and constraints in a high-level (symbolic) language. By blending data-driven and symbolic Artificial Intelligence (AI), the resulting conceptual hybrid SR/ASP framework ensures that the SR-generated symbolic expressions are not only statistically accurate, but also physically plausible, adhering to domain-specific constraints and principles, encoded into ASP (see Figure~\ref{fig_sr_asp}). This integration offers a promising direction for building versatile and trustworthy AI systems, capitalizing on the complementary strengths of learning from data and reasoning with knowledge \cite{platzer24}. It is worth emphasizing that, while ASP has been extensively applied across a wide range of domains \cite{erdem16,falkner18}, the integration of ASP with SR represents, to the best of our knowledge, the first effort to combine these two powerful frameworks.

On the whole, the present study highlights two major contributions: First, the ability of SR to translate intricate flow dynamics into simple, interpretable equations that effectively balance precision and clarity; and second, the significant role of knowledge-representation techniques in improving the reliability and domain-specific validity of data-driven SR models. These developments open new avenues for incorporating hybrid (data-driven and symbolic) methodologies into efficient computational systems, particularly in high-stakes fluid-dynamics scenarios where the integration of detailed, explainable simulations with real-time, data-driven insights is crucial.

The remainder of the article is organized as follows: The subsequent section (Section~\ref{section_flow_problem}) sets the mathematical formulation of the fluid-flow problems under investigation. Thereafter, Section~\ref{section_sr_model} introduces the architecture and parameters of the developed SR models. Section~\ref{section_results} is dedicated to the presentation and discussion of the results of the SR models. Section~\ref{section_sr_asp} presents the alluded hybrid SR/ASP approach. The article closes with a concluding section that summarizes the overall contribution and discusses interesting avenues for future research.

\section[Mathematical Formulation of the Fluid-Flow Problem]{Mathematical Formulation of the Fluid-Flow Problem}
\label{section_flow_problem}

The governing equations describing (time-independent) fluid flow in three spatial dimensions, without external forces, are the Navier-Stokes equations,

\vspace{2mm}

\begin{equation}\label{NavierStokesequations}
    \begin{array}{l}
        \rho\left(u'\frac{\partial u'}{\partial x'}+v'\frac{\partial u'}{\partial y'}+w'\frac{\partial u'}{\partial z'}\right)
        = -\frac{\partial p'}{\partial x'} + \mu\left(\frac{\partial^2 u'}{\partial {x'}^2}+\frac{\partial^2 u'}{\partial {y'}^2}+\frac{\partial^2 u'}{\partial {z'}^2}\right),\\[1ex]
        
        \rho\left(u'\frac{\partial v'}{\partial x'}+v'\frac{\partial v'}{\partial y'}+w'\frac{\partial v'}{\partial z'}\right)
        = -\frac{\partial p'}{\partial y'} + \mu\left(\frac{\partial^2 v'}{\partial {x'}^2}+\frac{\partial^2 v'}{\partial {y'}^2}+\frac{\partial^2 v'}{\partial {z'}^2}\right),\\[1ex]
        
        \rho\left(u'\frac{\partial w'}{\partial x'}+v'\frac{\partial w'}{\partial y'}+w'\frac{\partial w'}{\partial z'}\right)
        = -\frac{\partial p'}{\partial z'} + \mu\left(\frac{\partial^2 w'}{\partial {x'}^2}+\frac{\partial^2 w'}{\partial {y'}^2}+\frac{\partial^2 w'}{\partial {z'}^2}\right),
    \end{array}
\end{equation}

\vspace{3mm}

\noindent complemented by the conservation of mass equation,

\begin{equation}\label{Conservationequation}
    \frac{\partial u'}{\partial x'}+\frac{\partial v'}{\partial y'}+\frac{\partial w'}{\partial z'}=0,
\end{equation}

\noindent where $u',v',w'$ are the dimensional velocity components and $p'$ is the dimensional fluid pressure. Additionally, $\rho$ is the fluid density and $\mu$ is the dynamic fluid viscosity \cite{munson2012fundamentals}. To derive the non-dimensional form of the governing equations, the following non-dimensional parameters are introduced:

\begin{equation}\label{non-dimensional_parameters}
x = \frac{x'}{h}, \quad y = \frac{y'}{h}, \quad z = \frac{z'}{h}, \quad
u = \frac{u'\,h\,\rho}{\mu}, \quad v = \frac{v'\,h\,\rho}{\mu}, \quad w = \frac{w'\,h\,\rho}{\mu}, \quad
p = \frac{p'\,h^2\,\rho}{\mu^2},
\end{equation}

\vspace{2mm}

\noindent where $h$ represents the characteristic height \cite{tzirtzilakis12004biomagnetic}.

By substituting the parameters defined in Equation~\eqref{non-dimensional_parameters} into the dimensional Navier-Stokes equations (Equation~\eqref{NavierStokesequations}), the governing equations are recast in non-dimensional form. Applying the chain rule, in conjunction with the conservation of mass equation, yields the following closed-form expressions for the non-dimensional governing equations:

\begin{equation}\label{nondimnseq}
    \begin{array}{l}
        \frac{\partial \left(u^2\right)}{\partial x} + \frac{\partial \left(uv\right)}{\partial y} + \frac{\partial \left(uw\right)}{\partial z}
        = -\frac{\partial p}{\partial x} + \frac{\partial^2 u}{\partial x^2} + \frac{\partial^2 u}{\partial y^2} + \frac{\partial^2 u}{\partial z^2},\\[1ex]
        
        \frac{\partial \left(vu\right)}{\partial x} + \frac{\partial \left(v^2\right)}{\partial y} + \frac{\partial \left(vw\right)}{\partial z}
        = -\frac{\partial p}{\partial y} + \frac{\partial^2 v}{\partial x^2} + \frac{\partial^2 v}{\partial y^2} + \frac{\partial^2 v}{\partial z^2},\\[1ex]
        
        \frac{\partial \left(wu\right)}{\partial x} + \frac{\partial \left(wv\right)}{\partial y} + \frac{\partial \left(w^2\right)}{\partial z}
        = -\frac{\partial p}{\partial z} + \frac{\partial^2 w}{\partial x^2} + \frac{\partial^2 w}{\partial y^2} + \frac{\partial^2 w}{\partial z^2}.
    \end{array}
\end{equation}

\vspace{3mm}

\noindent It should be noted that applying the non-dimensional parameters of Equation~\eqref{non-dimensional_parameters} leads to non-dimensional equations that do not explicitly include non-dimensional groups, such as the Reynolds number $\mathrm{Re}$. In the present setting, the flow is instead driven by a prescribed constant streamwise pressure gradient $c$, which plays the role of the primary control parameter. A corresponding Reynolds number will be introduced later in this section as a convenient, equivalent parametrization of the operating condition.

The Finite Volume Method (FVM) is employed to discretize Equations~\eqref{nondimnseq} and the conservation of mass equation. The FVM discretizes the computational domain into smaller, finite control volumes, called cells \cite{mazumder2015numerical}. By integrating the governing equations over these control volumes, the system of non-linear partial differential equations is transformed into a larger system of non-linear {\em algebraic} equations, subject to corresponding boundary conditions. This transformation allows the equations to be solved using methods for non-linear systems, such as Newton's method. Additionally, the FVM evaluates equations at each cell centroid, enabling the implementation of various boundary conditions \cite{moukalled16}. 

The algebraic system obtained by the FVM discretization can be written in the form
\begin{equation}\label{Descretized}\small
    \begin{gathered}
        \frac{1}{2}\Delta y \Delta z\,(u_{E}^{2}-u_{W}^{2})
        +\frac{1}{2}\Delta x \Delta z\,(u_{N}v_{N}-u_{S}v_{S})
        +\frac{1}{2}\Delta x \Delta y\,(u_{T}w_{T}-u_{B}w_{B})
        +\Delta y \Delta z\,(p_{E}-p_{P}) \\
        -\frac{\Delta y \Delta z}{\Delta x}\,(u_{E}-2 u_{P}+u_{W})
        -\frac{\Delta x \Delta z}{\Delta y}\,(u_{N}-2 u_{P}+u_{S})
        -\frac{\Delta x \Delta y}{\Delta z}\,(u_{T}-2 u_{P}+u_{B}) = 0, \\[2mm]
        \frac{1}{2}\Delta y \Delta z\,(u_{E}v_{E}-u_{W}v_{W})
        +\frac{1}{2}\Delta x \Delta z\,(v_{N}^{2}-v_{S}^{2})
        +\frac{1}{2}\Delta x \Delta y\,(v_{T}w_{T}-v_{B}w_{B})
        +\Delta y \Delta z\,(p_{N}-p_{P}) \\
        -\frac{\Delta y \Delta z}{\Delta x}\,(v_{E}-2 v_{P}+v_{W})
        -\frac{\Delta x \Delta z}{\Delta y}\,(v_{N}-2 v_{P}+v_{S})
        -\frac{\Delta x \Delta y}{\Delta z}\,(v_{T}-2 v_{P}+v_{B}) = 0, \\[2mm]
        \frac{1}{2}\Delta y \Delta z\,(u_{E}w_{E}-u_{W}w_{W})
        +\frac{1}{2}\Delta x \Delta z\,(v_{N}w_{N}-v_{S}w_{S})
        +\frac{1}{2}\Delta x \Delta y\,(w_{T}^{2}-w_{B}^{2})
        +\Delta y \Delta z\,(p_{T}-p_{P}) \\
        -\frac{\Delta y \Delta z}{\Delta x}\,(w_{E}-2 w_{P}+w_{W})
        -\frac{\Delta x \Delta z}{\Delta y}\,(w_{N}-2 w_{P}+w_{S})
        -\frac{\Delta x \Delta y}{\Delta z}\,(w_{T}-2 w_{P}+w_{B}) = 0.
    \end{gathered}
\end{equation}
Here, \(\Delta x\), \(\Delta y\), and \(\Delta z\) denote the grid spacing in each coordinate direction, respectively. The velocity components $u_i$, $v_i$, $w_i$, and the pressure $p_{i}$ are evaluated at the control-volume faces \mbox{\(i \in \big\{E,W,N,S,T,B\big\}\)} and at the cell center \(P\). Additional details regarding the grid structure and control-volume arrangement can be found in the works \cite{chrimatopoulos2023,aravanis24b,katsoudas2025}.

The numerical solution of the system of Equation~\eqref{Descretized} is obtained using an in-house code implemented in \textsc{MATLAB}. The resulting algebraic system is solved on a collocated grid, using Newton's method combined with trust-region techniques. The solver settings include a structured grid of \(50 \times 25 \times 25\) control volumes, resulting in \(4 \times 50 \times 25 \times 25\) degrees of freedom (four unknowns \(u,v,w,p\) per control volume). Convergence is declared when the Euclidean norm of the residual vector falls below \(10^{-7}\).

By assuming that the velocity components \(v\) and \(w\) are identically zero throughout the flow, Equations~\eqref{nondimnseq} simplify significantly. In particular, one obtains \(\partial p / \partial x = c\), implying a \emph{linear} pressure drop along the channel length, and
\[
\frac{\partial^2 u}{\partial y^2} + \frac{\partial^2 u}{\partial z^2} = c,
\]
which yields a velocity profile \(\tilde{u}\) that depends on the imposed pressure gradient \(c\). This profile \(\tilde{u}\) is numerically evaluated for various values of \(c\), and is subsequently used as the inlet axial-velocity distribution.

The discrete system of Equation~\eqref{Descretized} is complemented by the following boundary conditions:
\begin{equation}\label{Boundary_conditions}
    \begin{gathered}
        \text{Inlet:}\quad 
        u = \tilde{u},\quad v = 0,\quad w = 0,\quad \frac{\partial p}{\partial x} = c, \\[1mm]
        \text{Outlet:}\quad 
        \frac{\partial u}{\partial x} = 0,\quad \frac{\partial v}{\partial x} = 0,\quad \frac{\partial w}{\partial x} = 0,\quad p = 0, \\[1mm]
        \text{Walls:}\quad 
        u = 0,\quad v = 0,\quad w = 0,\quad \frac{\partial p}{\partial y} = 0,\quad \frac{\partial p}{\partial z} = 0.
    \end{gathered}
\end{equation}
These conditions imply that the axial velocity component \(u\) follows the inlet profile \(\tilde{u}\), while the transverse components \(v\) and \(w\) are identically zero at the inlet. A constant streamwise pressure gradient \(c\) is imposed at the channel inlet. At the outlet, homogeneous Neumann boundary conditions are applied to all velocity components, whereas the pressure is fixed to zero. Along the channel walls, no-slip boundary conditions are enforced for the velocity components, and homogeneous Neumann boundary conditions are used for the pressure.

\begin{figure}
\centering
\includegraphics[scale=0.55]{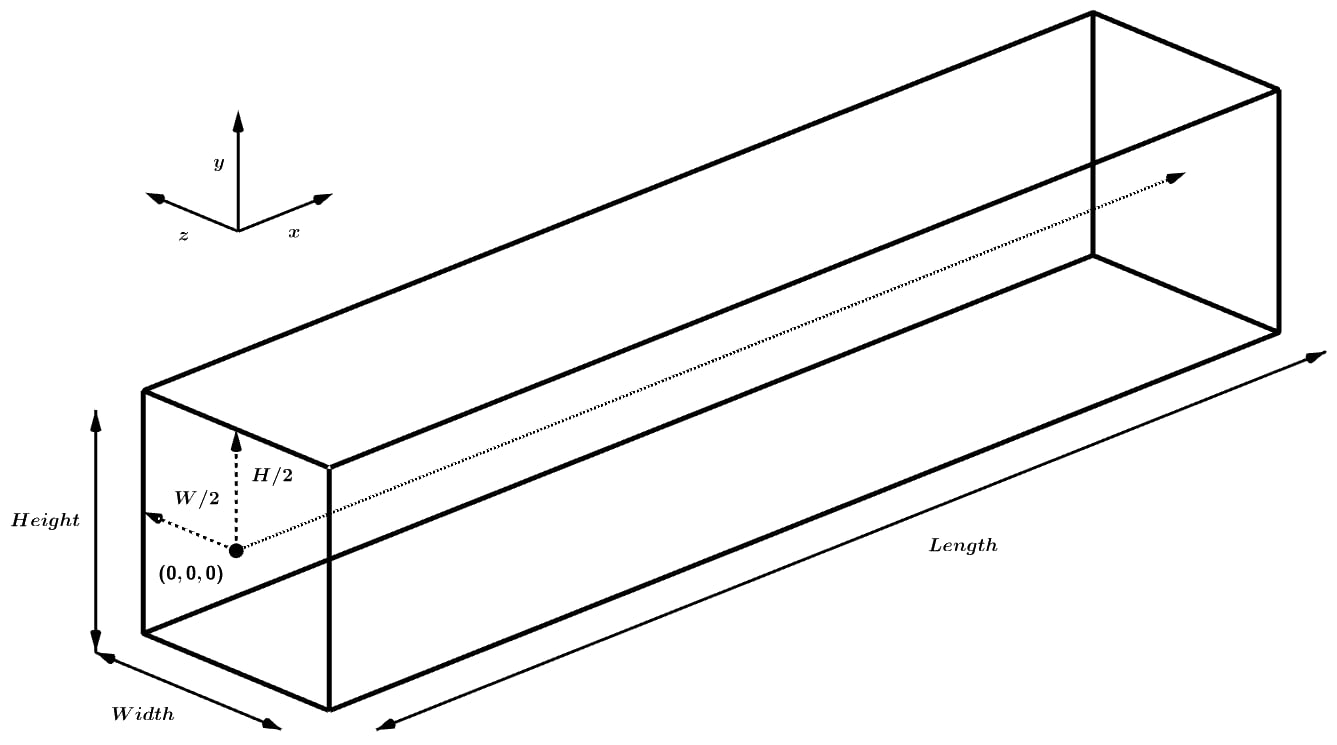}
\caption{Geometry of the three-dimensional symmetrical rectangular duct. The computational domain spans \mbox{$(0,L)\times(-H/2,H/2)\times(-W/2,W/2)$}. The duct height $H$ is used as the characteristic length scale.}
\label{3d_duct}
\end{figure}

In the present study, the fluid-mechanics problem under consideration is the fluid flow in a three-dimensional, symmetrical rectangular duct, with dimensions \mbox{$\left(0,L\right)\times\left(-H/2,H/2\right)\times\left(-W/2,W/2\right)$}. A visual representation of the geometry of the problem is shown in Figure~\ref{3d_duct}. Note that the characteristic length $h$ appearing in the identities of Equation~\eqref{non-dimensional_parameters} is assumed to be the height of the duct $H$; i.e., $h=H$. It is also noteworthy that, although the geometry of the problem appears relatively simple, deriving a corresponding analytical solution can be challenging. For example, Kaka\c{c} {\em et al.} \cite[Chapter~3]{kakac87} present an analytical solution expressed as an infinite series, underscoring the complexity of the problem, while also providing approximations that reduce computational costs.

To address the studied problem, numerical data were generated to train and test the Symbolic-Regression models by varying the driving pressure gradient. Specifically, a constant streamwise pressure gradient ranging from $c=-1000$ to $c=-8000$, with a step size of $-1000$, was applied to the Poisson equation, resulting in distinct fully developed inlet velocity profiles. The duct dimensions are specified as $L = 5$ and $H = W = 1$, and the fluid properties are kept fixed throughout.

As previously mentioned, the non-dimensional parameters of Equation~\eqref{non-dimensional_parameters} yield governing equations that do not explicitly include the Reynolds number. In this work, the flow regime is therefore most naturally indexed by the imposed pressure gradient $c$. For convenience of interpretation, we additionally compute an effective Reynolds number $\mathrm{Re}$ \emph{a posteriori} from the numerical solution $u$ via
\[
\mathrm{Re} = \int_{-W/2}^{+W/2} \int_{-H/2}^{+H/2} u \, dy \, dz,
\]
following Tzirtzilakis and colleagues \cite{tzirtzilakis12004biomagnetic}. For a fixed geometry and fluid, the laminar regime considered here ensures a one-to-one, monotonic relationship between the imposed pressure gradient $c$, the characteristic velocity, and the corresponding Reynolds number. Thus, $\mathrm{Re}$ functions as a \emph{case-level control parameter}, equivalent to having prescribed $\mathrm{Re}$ directly for each simulation. It is therefore not a per-sample function of the target field, but a global descriptor of the flow regime that labels each simulation (i.e., each value of $c$). 

The resulting {\em training dataset} includes the velocity components $u$, $v$, $w$, and the pressure $p$ within the 3D channel for operating conditions characterized by \mbox{$\mathrm{Re} \in \big\{34, 105, 174, 209, 279\big\}$}, while the {\em testing dataset} comprises the same quantities for $\mathrm{Re} \in \big\{70, 139, 244\big\}$. Consequently, the SR models are evaluated under {\em interpolation} scenarios with respect to this scalar control parameter. It is noted that the forthcoming analysis excludes the velocity components $v$ and $w$, as their magnitudes remain negligible and approach zero throughout the channel, as indicated by the numerical solutions.

\section{The Symbolic-Regression Models}
\label{section_sr_model}

Symbolic Regression (SR) is a powerful computational technique used to uncover interpretable mathematical relationships that govern data dynamics. Unlike traditional regression methods that rely on a fixed, predefined functional form, SR explores a flexible search space constructed from user-specified mathematical primitives to derive models that are both accurate and interpretable. This flexibility makes SR particularly suited for domains where the underlying physical relationships are unknown or complex.

SR is typically implemented using evolutionary algorithms, which draw inspiration from biological processes like natural selection and genetic mutation \cite{petrowski17}. These algorithms treat mathematical expressions as individuals within a population, and evolve them over successive generations to optimize their performance against a fitness criterion, such as minimizing prediction error. The process begins with the generation of an initial population of random mathematical expressions, composed of variables, constants, and operators. Each of these expressions is then evaluated for fitness, based on how accurately it approximates the observed data using predefined metrics like Mean Squared Error or custom loss functions. The population is subsequently evolved by introducing variation through operations such as mutation, which randomly alters parts of an expression, and crossover, which combines segments from two parent expressions to produce new candidates. Finally, a selection and refinement step retains the most accurate and interpretable expressions, gradually improving the population across generations. This biologically inspired work-flow enables SR to efficiently explore a vast space of candidate mathematical relationships, striking a balance between predictive accuracy and interpretability. The outcome is a set of concise, human-readable equations that capture the underlying dynamics of the system being modelled.

For implementing SR in the present study, the \texttt{PySR} tool (version 0.19.4) was employed. \texttt{PySR} is a powerful SR library written in Python and Julia, leveraging techniques such as regularized evolution, simulated annealing, and gradient-free optimization \cite{cranmer23}. In the remainder of this section, we detail the design and implementation of the SR models, highlighting their configuration and providing the rationale behind each methodological choice.

To begin with, two distinct SR models were developed: One specifically designed to predict the axial velocity $u$, and the other to model the pressure $p$ within the 3D channel. In constructing these models, interpretability was encouraged by restricting the maximum allowed expression complexity, and by selecting a domain-motivated set of primitive mathematical operators (e.g., arithmetic operations and low-order polynomial functions). These primitive operators also constrain the search space, thereby preserving computational tractability.

More specifically, the models were trained for $1000$ iterations. This setting was supported by trial runs using markedly larger iteration budgets, which indicated that the resulting expressions and error metrics remained effectively unchanged. Core arithmetic operations, such as addition ($+$), subtraction ($-$), multiplication ($\ast$), and division ($\div$), formed the basis for constructing expressions. To capture more complex relationships, non-linear operators, including \texttt{square} and \texttt{cube} functions, were also incorporated as ``building blocks''. This choice is motivated by fluid-mechanics principles, which indicate that laminar fluid-flow solutions frequently exhibit polynomial-like behaviour --- allowing velocity and pressure profiles to be well-approximated by low-order polynomials, as will be further discussed in Subsection~\ref{subsection_analytical}, where we examine relevant {\em analytical} equations from the literature. It should be stressed that restricting the operator set in a domain-informed manner is standard practice in SR. For example, in the seminal \texttt{PySR} paper, Cranmer notes that ``empirical relations are frequently comprised of operators which are unique to one particular field of science, so the equation search strategy must allow for custom operators'' \cite[p.~3]{cranmer23}, underscoring the importance of incorporating domain-specific, user-defined operators within \texttt{PySR}.

To enhance the simplicity and utility of the resulting equations, a number of constraints were introduced. The {\em complexity} of expressions ---expressed by the total count of mathematical operations and terms within each equation--- was limited by setting a maximum size ($25$), ensuring they remained interpretable. Additionally, nested constraints were applied to prevent unnecessary combinations of higher-order functions, such as squaring a squared term. This helped reduce redundancy and promoted concise mathematical representations. Reproducibility was ensured by explicitly configuring \texttt{PySR} in deterministic mode, using a fixed \texttt{random_state} and disabling parallelism so that repeated runs yield identical results.

As outlined in Section~\ref{section_flow_problem}, the training and testing datasets comprise the spatial coordinates ($x$, $y$, $z$) together with the Reynolds number ($\mathrm{Re}$) as input variables, while the corresponding outputs are the axial velocity ($u$) and the pressure ($p$). Here, $\mathrm{Re}$ plays the role of a scalar control parameter that indexes the operating condition of each simulation (through the imposed pressure gradient), since all samples originating from the same numerical run share a {\em single} $\mathrm{Re}$ value. The SR models therefore learn a {\em parametric} mapping of the form $u = u(x,y,z;\mathrm{Re})$ and $p = p(x,y,z;\mathrm{Re})$ across different flow regimes, rather than a self-referential dependence of $u$ on its own pointwise values. For clarity, the upper-case variables $X$, $Y$, $Z$ appearing in the forthcoming SR equations denote the same non-dimensional coordinates as $x$, $y$, $z$ used in the mathematical formulation of the fluid-flow problem. A summary of the key parameters and settings employed by the \texttt{PySR} tool is provided in Table~\ref{table_sr_parameters}.

\begin{table}[t]
\centering\footnotesize\renewcommand{\arraystretch}{1.5}
\begin{tabular}{|M{3.5cm}|M{3.5cm}|M{6.5cm}|}
\hline
\textbf{Parameter} & \textbf{Value} & \textbf{Description} \\ \hline\hline
\texttt{niterations} & $1000$ & Number of training iterations \\ \hline
\texttt{binary\_operators} & $+$ , $-$ , $*$ , $/$ & Core arithmetic operations \\ \hline
\texttt{unary\_operators} & \texttt{square(x) = x²} , \texttt{cube(x) = x³} & Low-order polynomials as building blocks \\ \hline
\texttt{maxsize} & $25$ & Limits expression complexity \\ \hline
\texttt{nested\_constraints} & \texttt{True} & Restricts nested \texttt{square} and/or \texttt{cube} operators \\ \hline
\texttt{variable\_names} & $X$ , $Y$ , $Z$ , $\mathrm{Re}$ & Names of input variables $x$ , $y$ , $z$ , $\mathrm{Re}$ \\ \hline
\end{tabular}
\caption{Key parameters of the \texttt{PySR} tool used for the development of the two SR models.}
\label{table_sr_parameters}
\end{table}

During the training process, the SR models iteratively refine their mathematical expressions, striving to minimize the prediction error quantified by a {\em loss function} (fitness criterion). \texttt{PySR} uses Mean Squared Error (MSE) as its default loss function to evaluate the performance of candidate equations, adhering to predefined configurations, as outlined in Table~\ref{table_sr_parameters}. For a target quantity \( y \), the MSE is defined as $\text{MSE} = \frac{1}{n} \displaystyle\sum_{i=1}^{n} \left( y_i - \hat{y}_i \right)^2$, where \(n\) is the number of data points, \(y_i\) represents the reference (actual) value of the \(i\)-th data point, and \(\hat{y}_i\) denotes the SR-predicted value for the \(i\)-th data point.

\begin{figure}[h!]
\centering\small
\scalebox{1}{
\begin{tikzpicture}[very thick]
    \draw[->, >=stealth] (0,0) -- (6,0) node[right] {Loss};
    \draw[->, >=stealth] (0,0) -- (0,6) node[above] {Complexity};
    \draw[green, domain=0.5:5.5, samples=50, smooth]
        plot (\x, {2.8/\x});
    \foreach \x in {0.6, 0.7, 1, 1.4, 1.9, 3, 3.8, 4.5, 5.4} {
        \fill[red] (\x, {2.8/\x}) circle (5pt); }
    \foreach \x/\y in {2/2, 3.3/1.5, 2/3, 4.8/2, 3/3, 4.3/4, 3.5/2.5, 2.5/4.5, 4.7/3.5, 1.5/4, 2/5} {
        \fill[gray!70] (\x, \y) circle (5pt);
    }
\end{tikzpicture}}
\caption{The Pareto front (green line) illustrating the optimal trade-off between model loss and complexity. Red markers denote models (equations) on the front, representing the best balance of accuracy and simplicity, while gray points indicate models (equations) that are less optimal in one or both aspects.}
\label{fig_pareto}
\end{figure}
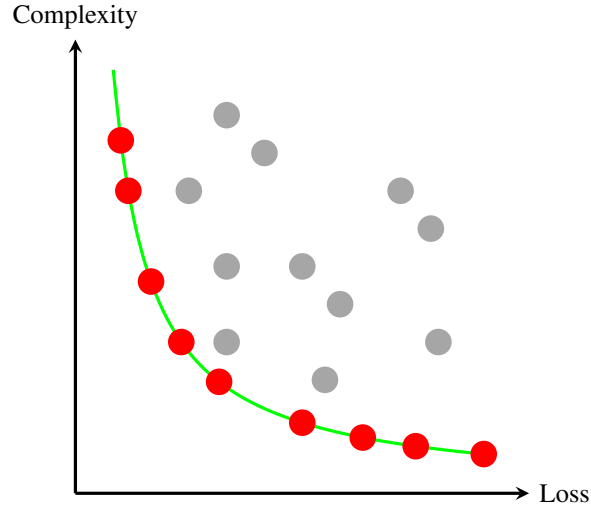

Upon completing the training process, \texttt{PySR} produces a collection of candidate models that form the {\em Pareto front} \cite{cranmer23}. In multi-objective optimization, the Pareto front is the set of all Pareto-efficient solutions, namely, the set of solutions that cannot be improved in any one objective without degrading at least one of the other objectives \cite{wang11}. In our context, the Pareto front represents the optimal trade-off between simplicity (model complexity) and accuracy (error metrics such as MSE or MAE). Formally, a model belongs to the Pareto front if and only if no other model achieves {\em both} a lower complexity and a lower error. Hence, models on the Pareto front are characterized by their {\em best balance} between interpretability and predictive accuracy, ensuring robustness and minimizing the risk of overfitting (see Figure~\ref{fig_pareto}).

\section{Results and Discussion}
\label{section_results}

Having discussed the architecture of our SR models, we turn to the presentation of the derived results concerning the axial velocity $u$ and the pressure $p$ of the channel.

The equations derived from the SR models are assessed using three primary error metrics: Mean Squared Error (MSE), Mean Absolute Error (MAE), and Normalized Mean Absolute Error (NMAE). The MSE, introduced in Section~\ref{section_sr_model}, serves as the default loss function (fitness criterion) employed by \texttt{PySR} to evaluate candidate expressions. Meanwhile, focusing on the prediction of a target quantity \( y \), MAE and NMAE are defined as $\text{MAE} = \frac{1}{n} \displaystyle\sum_{i=1}^{n} |y_i - \hat{y}_i|$ and $\text{NMAE} = \frac{\text{MAE}}{\max(y)-\min(y)}\, 100\%$, where \( n \) is the total number of data points, \( y_i \) is the reference (actual) value of the $i$-th data point, \( \hat{y}_i \) is the SR-predicted value for the $i$-th data point, and $\max(y)$, $\min(y)$ are the maximum and minimum reference values of $y$, respectively.

\subsection{Symbolic Equations}
\label{subsection_equations}

Against this background, two symbolic equations derived from the implemented SR models are presented below. These expressions, selected from the Pareto front of their respective models, exemplify an optimal balance between accuracy and complexity, positioning them as robust and interpretable candidates for further exploration and analysis. The computation time required for an SR model to generate the full set of Pareto-front equations is approximately 15 minutes on a system equipped with a 13th-generation Intel\textsuperscript{\textregistered} Core\texttrademark{} i7--13700 processor (2.10~GHz) and \(32\)~GB of RAM.

\vspace{3mm}

\begin{center}
\begin{tcolorbox}[colback=white!90!gray, colframe=black, sharp corners, width=0.6\textwidth, box align=top, valign=top]
\begin{align}
u^{\text{symb}} &= \mathrm{Re}\, \Big( 2.18 - 8.46\, Y^2 \Big)\Big( 1 - 3.89\, Z^2 \Big) \label{eq_u} \\[0.5cm]
p^{\text{symb}} &= \mathrm{Re}\, \Big( 143.43 - 28.69\, X \Big) \label{eq_p}
\end{align}
\vspace{-8pt}
\end{tcolorbox}
\end{center}

\vspace{3mm}

Equation~\eqref{eq_u} defines the axial velocity \( u \) within the channel as a function of the Reynolds number \( \mathrm{Re} \) and the spatial variables \( Y \) and \( Z \). The expression incorporates quadratic dependencies on both \( Y \) and \( Z \), with coefficients that capture the influence of \( \mathrm{Re} \) on the velocity distribution. The quadratic terms establish a {\em parabolic} velocity profile, consistent with the expected flow characteristics in such a configuration. The pressure field \( p \) is represented symbolically in Equation~\eqref{eq_p} as a function of the Reynolds number \( \mathrm{Re} \) and the spatial variable \( X \). This equation reveals a negative linear relationship between \( p \) and \( X \), where the coefficients are scaled by \( \mathrm{Re} \). The negative correlation between \( p \) and \( X \) highlights a linear pressure drop across the channel, a common feature in laminar flow systems.

Table~\ref{table_equations} provides a detailed comparison of the symbolic expressions, highlighting both their complexity and performance on the training and testing datasets. The performance is evaluated using the error metrics MSE, MAE, and NMAE, quantifying the accuracy of the symbolic models for the axial velocity (\( u^{\mathrm{symb}} \)) and the pressure field (\( p^{\mathrm{symb}} \)), with results reported separately for the training and testing datasets. As already stated, the complexity of each equation, shown in the second column of Table~\ref{table_equations}, reflects the total count of mathematical operations and terms within each expression. A higher complexity value suggests a more intricate equation, which often results from the need to capture detailed relationships in the data.

\vspace{2mm}

\begin{table}[h!]
\centering\footnotesize
\renewcommand{\arraystretch}{1.7}
\setlength{\tabcolsep}{8pt}
\begin{tabular}{|l|c|ccc|ccc|}
\hline
\multirow{2}{*}{\textbf{Equation}} & \multirow{2}{*}{\textbf{Complexity}} & \multicolumn{3}{c|}{\textbf{Training Dataset (62.5\%)}} & \multicolumn{3}{c|}{\textbf{Testing Dataset (37.5\%)}} \\ \cline{3-8}
 &  & \textbf{MSE} & \textbf{MAE} & \textbf{NMAE} & \textbf{MSE} & \textbf{MAE} & \textbf{NMAE} \\ \hline\hline
$u^{\mathrm{symb}}$ \eqref{eq_u} & $17$ & $44.98$ & $4.81$ & $0.008 \%$ & $47.62$ & $5.36$ & $0.010 \%$ \\ \hline\hline
$p^{\mathrm{symb}}$ \eqref{eq_p} & $9$ & $1.68\cdot 10^{-6}$ & $0.0011$ & $2.75\cdot 10^{-8} \%$ & $1.8 \cdot 10^{-6}$ & $0.0012$ & $3.44\cdot 10^{-8} \%$ \\ \hline
\end{tabular}
\caption{Measures of accuracy for Equations~\eqref{eq_u} and~\eqref{eq_p} of the SR models.}
\label{table_equations}
\end{table}

For \( u^{\mathrm{symb}} \) (Equation~\eqref{eq_u}), with a complexity of $17$, the model achieves a low NMAE of \( 0.008 \% \) on the training dataset and \( 0.010 \% \) on the testing dataset. These results demonstrate a highly accurate fit, with slightly better performance observed during training. The pressure field, \( p^{\mathrm{symb}} \) (Equation~\eqref{eq_p}), has a notably lower complexity of $9$ but achieves remarkable precision. The MSE values for both the training and testing datasets are in the order of \( 10^{-6} \), while the NMAE is consistently in the range of \( 10^{-8} \% \). This demonstrates that the symbolic model for \( p \) not only captures the underlying trend but does so with minimal error, even for unseen data.

Overall, the results highlight the balance between complexity and accuracy achieved by the SR approach. The models effectively generalize from the training data to the testing data, with minimal loss in accuracy, showcasing their robustness and suitability for modelling fluid velocity and pressure within the channel. The consistent performance across the training and testing datasets suggests that overfitting was effectively mitigated, reinforcing the reliability of the approach.

\subsection{Representative Plots}
\label{subsection_plots}

\begin{figure}[h!]
\centering
\captionsetup[subfigure]{labelformat=empty, justification=centering}
\begin{subfigure}[t]{0.48\textwidth}
    \centering
    \includegraphics[width=\textwidth]{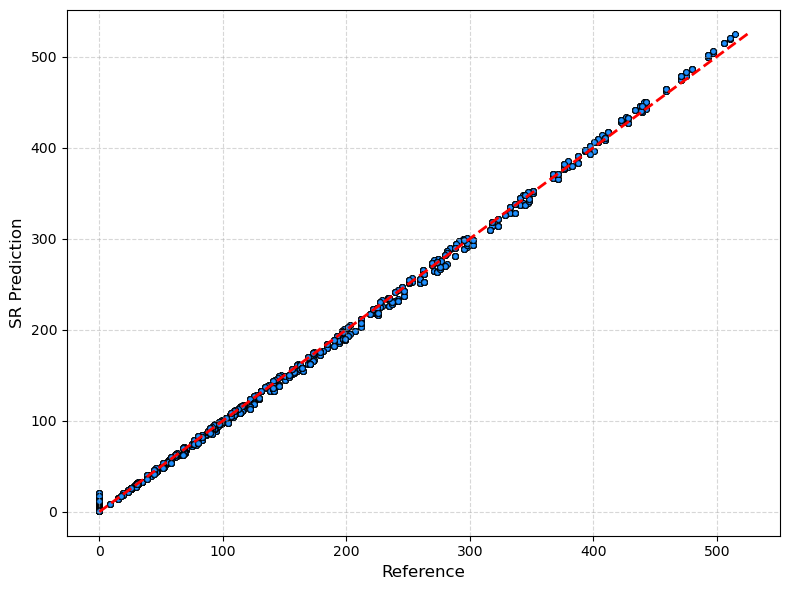}
    \caption{\small Axial velocity $u$: Equation~\eqref{eq_u}}
\end{subfigure}
\hfill
\begin{subfigure}[t]{0.48\textwidth}
    \centering
    \includegraphics[width=\textwidth]{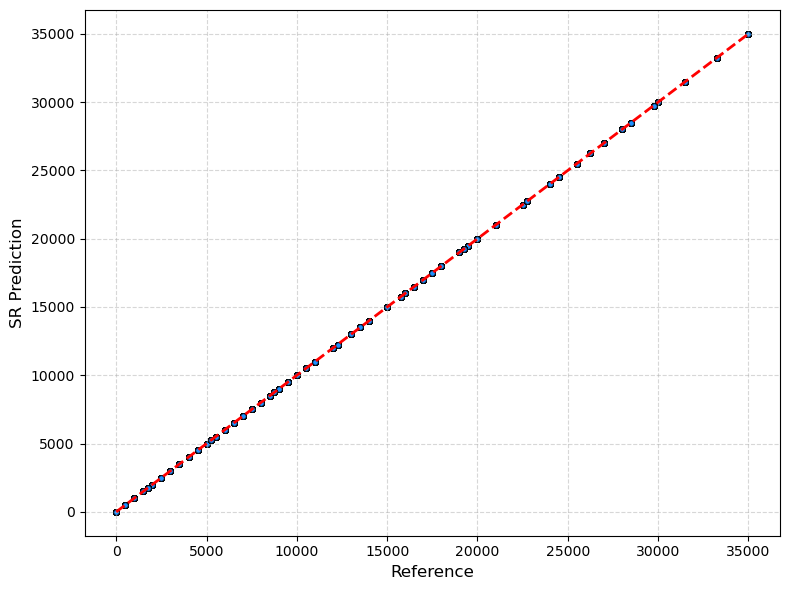}
    \caption{\small Pressure $p$: Equation~\eqref{eq_p}}
\end{subfigure}
\caption{Identity plots for Equations~\eqref{eq_u} and~\eqref{eq_p} of the SR models. The red $45^\circ$ line represents a perfect match between the symbolic equation derived from an SR model and the numerical results.}
\label{fig_identity}
\end{figure}

The low error metrics presented in Table~\ref{table_equations} are further substantiated by the exceptional alignment observed in the fitted {\em identity} plots for Equations~\eqref{eq_u} and~\eqref{eq_p}, which compare the predictions of the SR models against the reference numerical results, as illustrated in Figure~\ref{fig_identity}.

\vspace{2mm}

\begin{figure}
\centering
\captionsetup[subfigure]{labelformat=empty, justification=centering}
\begin{subfigure}[t]{0.72\textwidth}
    \centering
    \includegraphics[width=\textwidth]{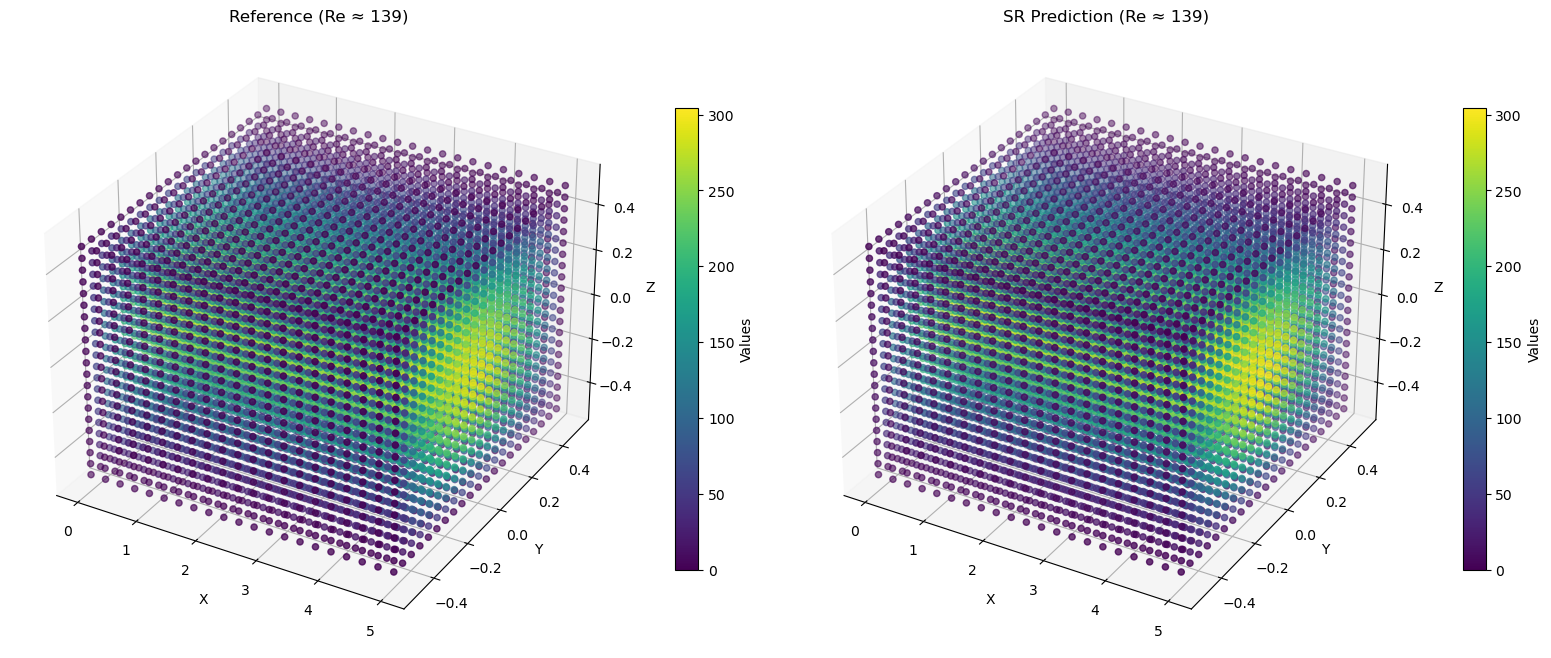}

    \hfill\vspace{1mm}

    \includegraphics[width=0.9\textwidth]{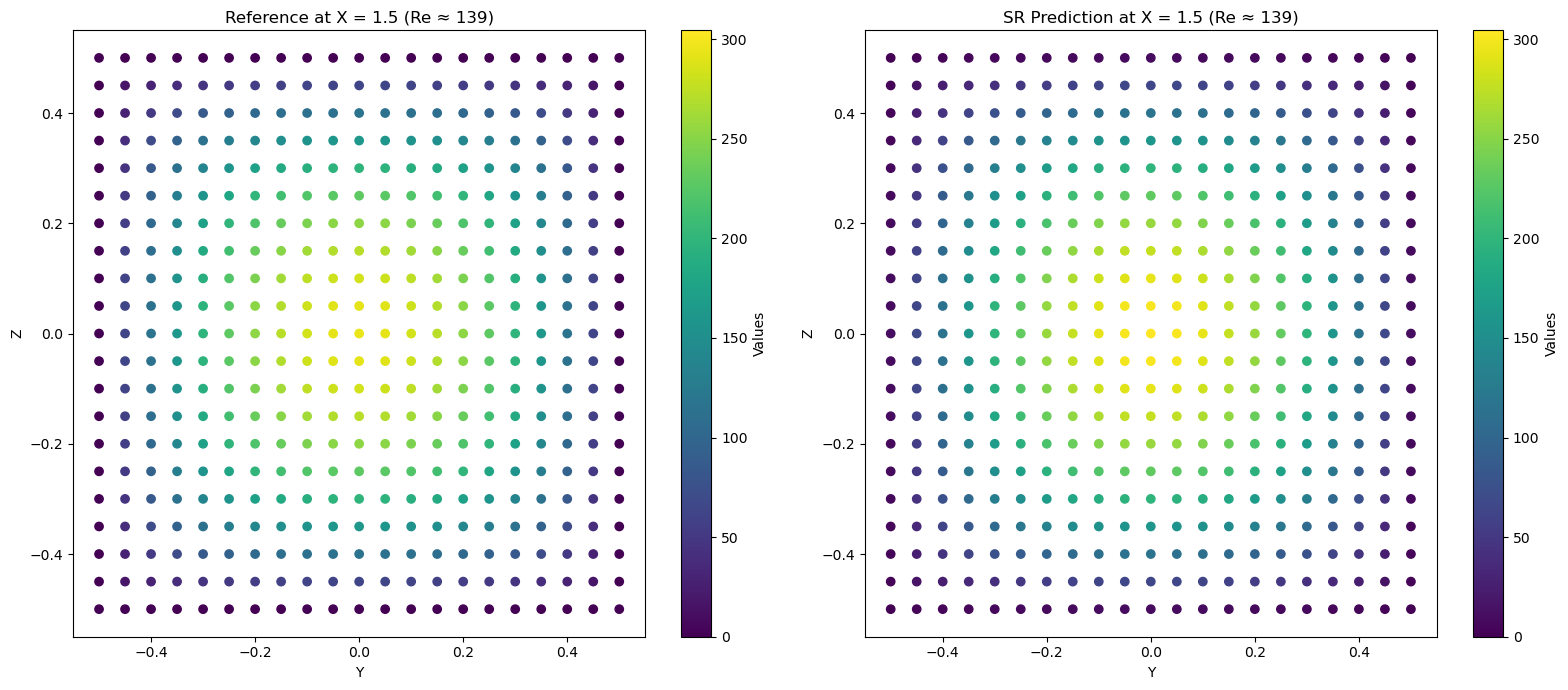}
    \caption{\footnotesize $u$ for $\mathrm{Re} \approx 139$: 3D contour-map (top) and 2D profile at $x = 1.5$ (bottom).}
\end{subfigure}

\hfill \vspace{8mm}

\begin{subfigure}[t]{0.72\textwidth}
    \centering
    \includegraphics[width=\textwidth]{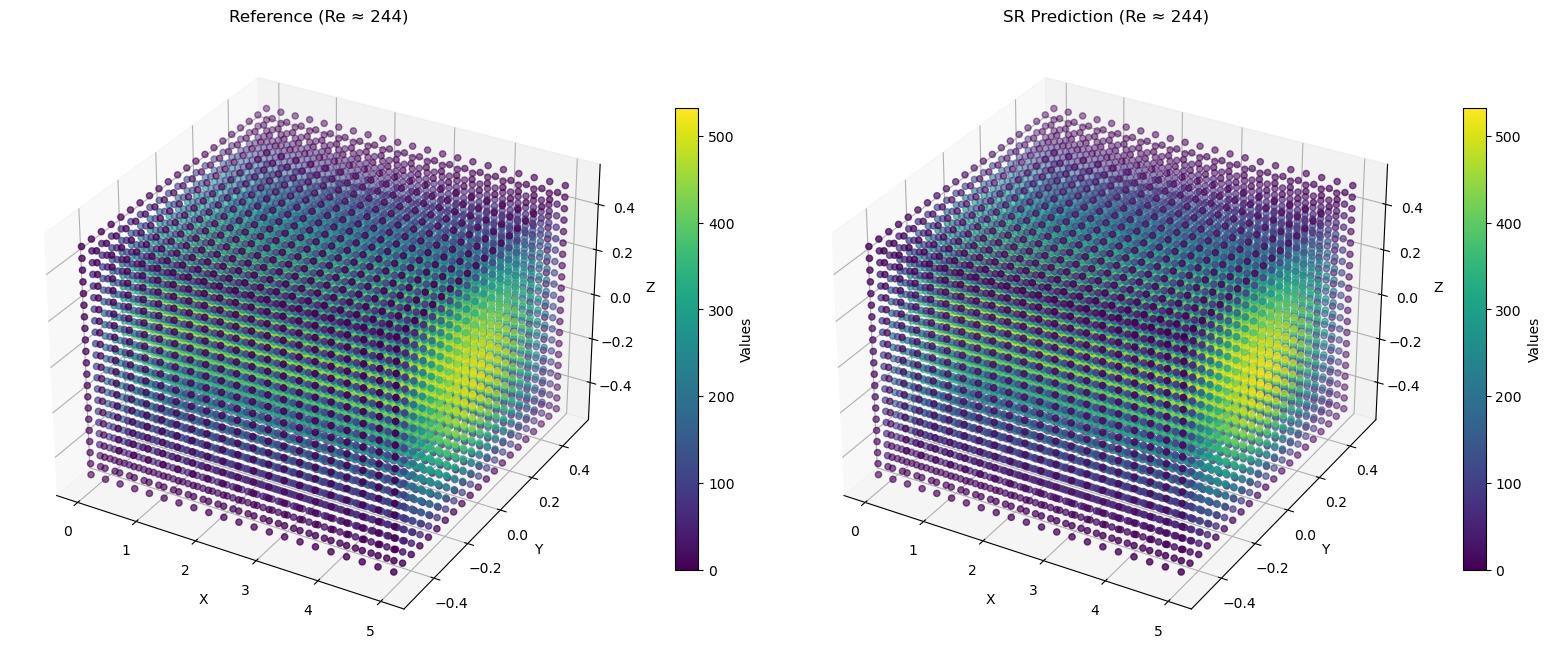}

    \hfill\vspace{1mm}

    \includegraphics[width=0.9\textwidth]{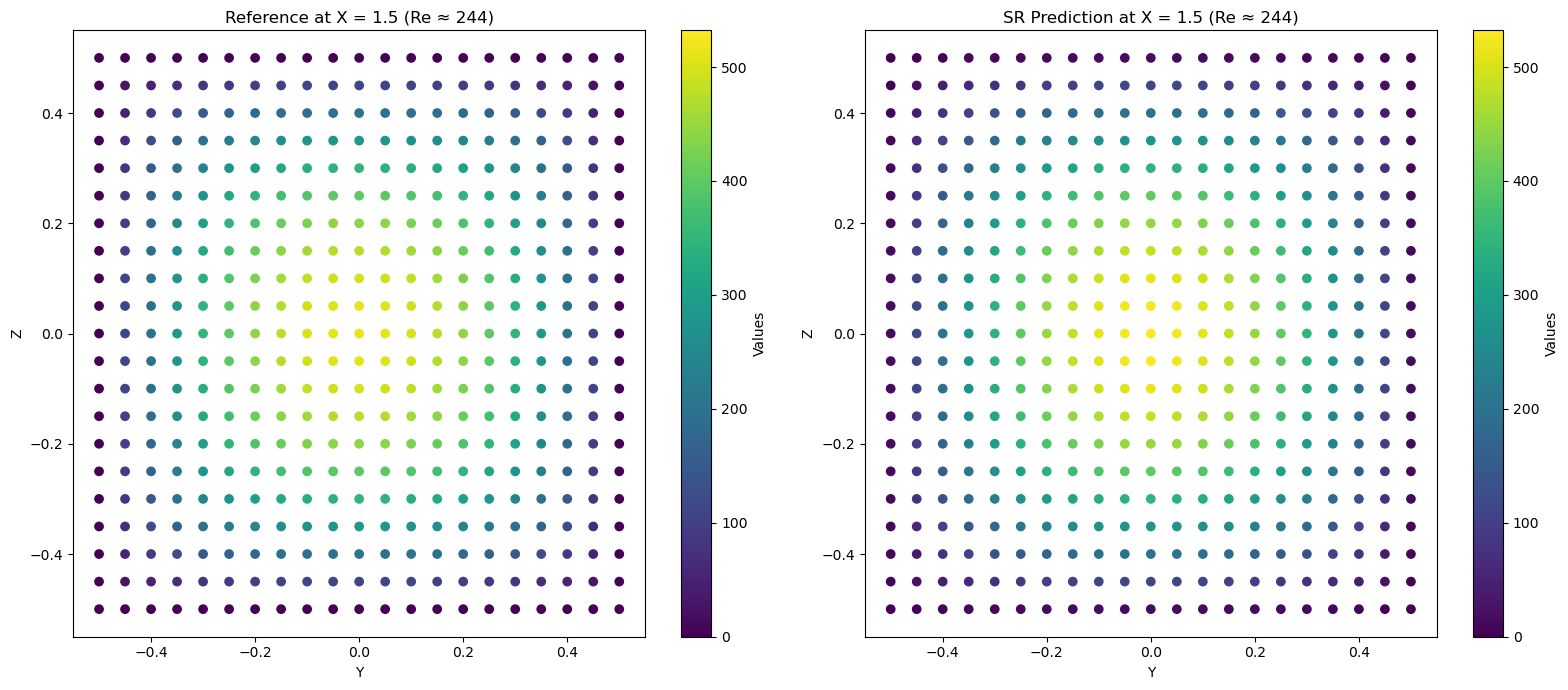}
    \caption{\footnotesize $u$ for $\mathrm{Re} \approx 244$: 3D contour-map (top) and 2D profile at $x = 1.5$ (bottom).}
\end{subfigure}
\caption{3D contour-maps of the axial velocity $u$ within the channel, alongside its 2D profile at a representative vertical cross-section of the channel, for indicative Reynolds numbers of the testing dataset. Depicted are the predicted quantities as generated by the SR model, and the reference quantities as generated by the numerical solutions.}
\label{fig_velocity}
\end{figure}

\begin{figure}[h!]
\centering
\captionsetup[subfigure]{labelformat=empty, justification=centering}
\begin{subfigure}[t]{\textwidth}
    \centering
    \includegraphics[width=\textwidth]{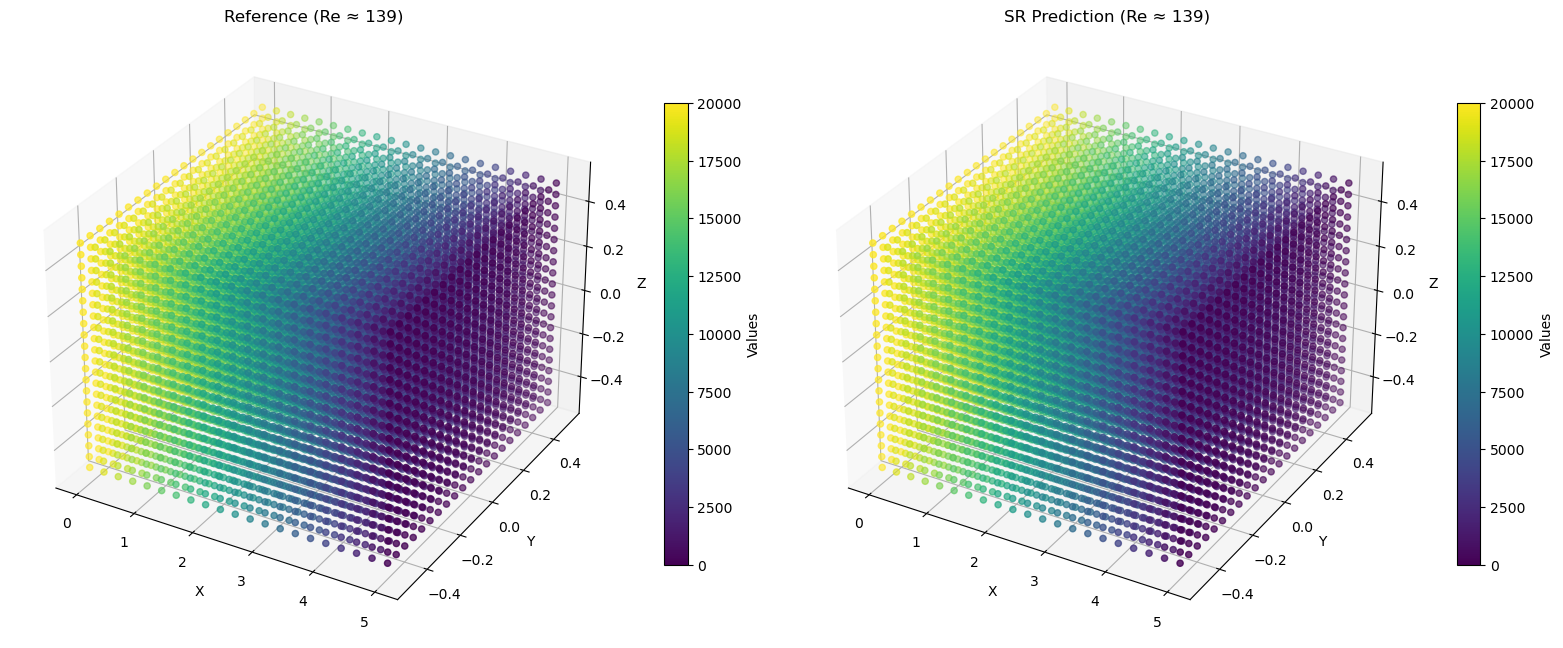}
    \caption{\small $\mathrm{Re} \approx 139$}
\end{subfigure}

\hfill \vspace{5mm}

\begin{subfigure}[t]{\textwidth}
    \centering
    \includegraphics[width=\textwidth]{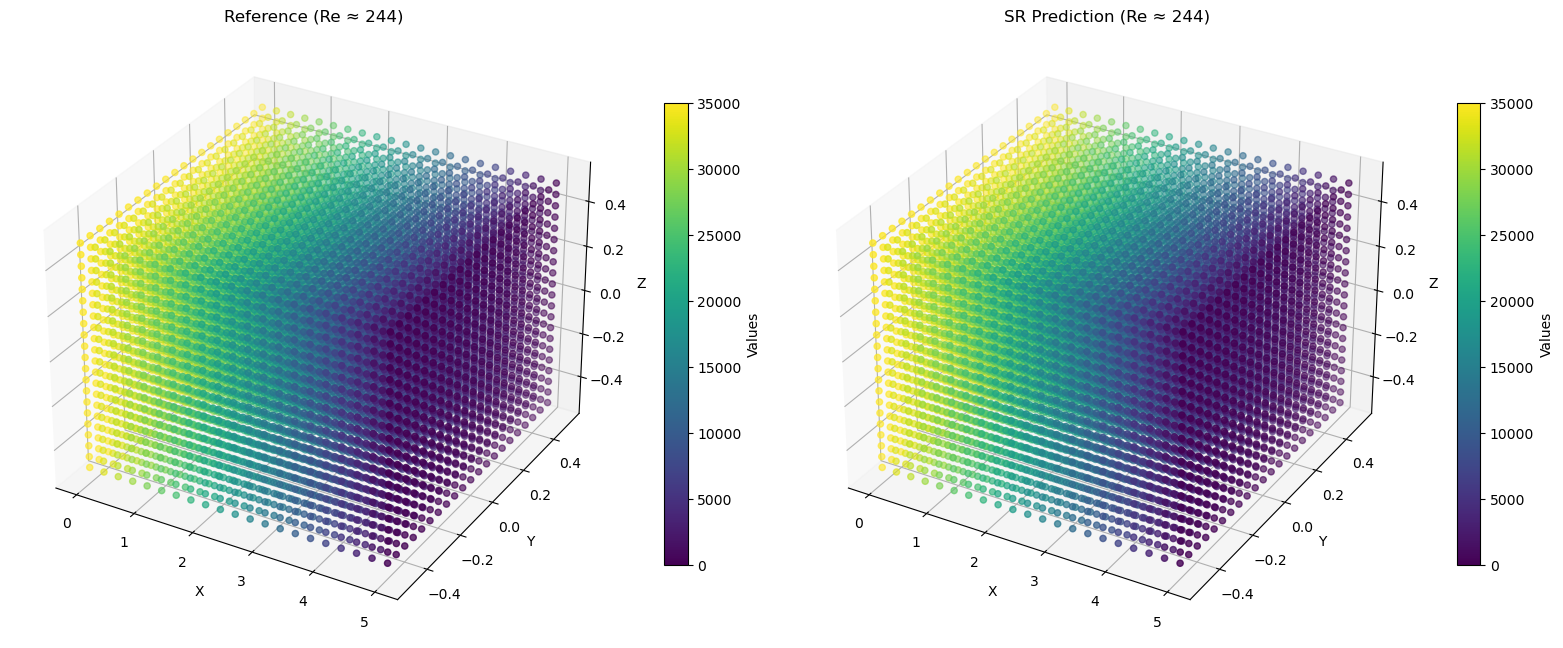}
    \caption{\small $\mathrm{Re} \approx 244$}
\end{subfigure}
\caption{3D contour-maps of the pressure $p$ within the channel, for indicative Reynolds numbers of the testing dataset. Depicted are the predicted quantities as generated by the SR model, and the reference quantities as generated by the numerical solutions.}
\label{fig_pressure}
\end{figure}

Furthermore, Figure~\ref{fig_velocity} depicts the 3D contour-maps of the axial velocity $u$ within the channel, alongside its 2D profile at a representative vertical cross-section of the channel, for indicative Reynolds numbers of the testing dataset. Depicted are the predicted quantities as generated by the SR model, and the reference quantities as generated by the numerical solutions. Along similar lines, Figure~\ref{fig_pressure} depicts the 3D contour-maps of the pressure $p$ within the channel, for indicative Reynolds numbers of the testing dataset. Depicted are the predicted quantities as generated by the SR model, and the reference quantities as generated by the numerical solutions.

Evidently, the SR models demonstrate exceptional accuracy and interpretability, not only in capturing the flow's axial velocity field, but also in representing its pressure distribution. For the axial velocity $u$, the derived equation produces nearly indistinguishable results from the reference numerical solution across the tested Reynolds numbers. The predicted flow fields preserve the quintessential parabolic profile, smoothly peaking at the channel's centerline and tapering off near the boundary walls in accordance with the no-slip boundary conditions. This fidelity to fundamental physical behaviour is quantified by exceedingly low NMAEs, consistently at the order of a few thousandths of a percent. Such negligible discrepancies underscore the model's remarkable capability to generalize beyond the training dataset and its promise for reducing reliance on high-fidelity, computationally expensive simulations. The SR model developed for the pressure provides an equally compelling representation of the pressure field. The retrieved symbolic expression accurately captures the expected linear pressure gradient along the channel's longitudinal direction, a hallmark feature of laminar, fully developed flow. The minimal residual errors in pressure predictions, as reflected by the near-vanishing error metrics, attest to the robustness of the model and its ability to encode fundamental fluid-dynamic principles in a compact functional form.

Taken together, the SR-derived relationships for both axial velocity and pressure distil complex physical phenomena into transparent, mathematically concise expressions, streamlining the pathway from data-driven insights to physically interpretable models.

\subsection{Literature's Analytical Equations}
\label{subsection_analytical}

Following the presentation of the concrete results obtained from the SR models, we now delve into {\em analytical} equations from the literature relevant to laminar fluid flows in 3D channels. A foundational example is the exit velocity profile of an incompressible fluid flowing steadily through a rectangular duct, as illustrated in Figure~\ref{3d_duct} of Section~\ref{section_flow_problem}. This profile is described approximately by the equation

\begin{equation} \label{eq_u_ref_general}
u = u_{\text{max}} \left( 1 - 4\,\frac{y^2}{H^2} \right) \left( 1 - 4\,\frac{z^2}{W^2} \right),
\end{equation}

\vspace{2mm}

\noindent where \(u\) is the axial velocity at any point within the cross-section, \(u_{\text{max}}\) is the maximum velocity at the center of the duct, \(H\) is the height of the duct in the \(y\)-direction, \(W\) is the width of the duct in the \(z\)-direction, and \(y\), \(z\) are the coordinates in the cross-sectional plane \cite[p. 198, Problem P3.18]{white11}. (For exact solutions of fully developed velocity profiles in flows through rectangular ducts, the reader is referred to the work of Kaka\c{c} {\em et al.} \cite{kakac87}.) In the context of our study, where the duct dimensions are defined as $H = W = 1$ (Section~\ref{section_flow_problem}), Equation~\eqref{eq_u_ref_general} simplifies to

\begin{equation} \label{eq_u_ref}
u = u_{\text{max}} \Big( 1 - 4\, y^2 \Big) \Big( 1 - 4\, z^2 \Big).
\end{equation}

\vspace{3mm}

The above expression highlights the {\em parabolic} velocity profile commonly observed in laminar flows through ducts, with $u_{\text{max}}$ at the center and the velocity decreasing symmetrically toward the walls, due to the no-slip boundary conditions.

Building on this foundation, the conducted numerical simulation suggests a direct relationship between the Reynolds number $\mathrm{Re}$ and the maximum axial velocity at the centerline of the channel $u_{\text{max}}$, given by

\begin{equation} \label{eq_re_u_max}
\mathrm{Re} \approx \frac{u_{\text{max}}}{2.109}.
\end{equation}

By combining Equations~\eqref{eq_u} and~\eqref{eq_re_u_max}, we obtain the following symbolic representation for the velocity profile:

\begin{equation} \label{eq_u_symb}
\begin{aligned}
u^{\text{symb}} &\approx \frac{u_{\text{max}}}{2.109} \Big( 2.18 - 8.46\, Y^2 \Big) \Big( 1 - 3.89\, Z^2 \Big) = u_{\text{max}} \Big( 1.03 - 4.01\, Y^2 \Big) \Big( 1 - 3.89\, Z^2 \Big).
\end{aligned}
\end{equation}

\vspace{3mm}

Evidently, the forms of Equations~\eqref{eq_u_ref} and~\eqref{eq_u_symb} are strikingly similar. The symbolic representation in Equation~\eqref{eq_u_symb}, derived from the corresponding SR model, provides an excellent approximation to the analytical solution from the literature, namely Equation~\eqref{eq_u_ref}. The small remaining discrepancy is primarily attributed to the finite-resolution numerical data. Similarly, the symbolic representation of the pressure field in Equation~\eqref{eq_p}, derived from the corresponding SR model, aligns closely with the general form of pressure distributions reported in the literature, according to which the pressure {\em drops linearly} across the length of the channel \cite{white11,pritchard11}. The observed remarkable similarity underscores the ability of the developed SR models to capture the fundamental characteristics of the velocity and pressure fields with high fidelity, reinforcing their potential as powerful tools for both theoretical studies and real-world engineering applications.

\subsection{Robustness Analysis}
\label{subsection_robustness}

Although our primary analysis relies on the deterministic dataset generated by the numerical simulation, in this subsection we additionally assess the robustness of the proposed approach by injecting controlled levels of synthetic {\em noise} into the existing data and retraining the SR models. To this end, we added 2\% noise to the axial velocity $u$ and the pressure $p$, leaving all other variables unchanged. Accordingly, two symbolic equations derived from the Pareto front of the noise-augmented SR models are presented below.

\vspace{-4mm}

\begin{center}
\begin{align}
u^{\text{symb\_noise}} &= \mathrm{Re}\, \Big( 2.17 - 8.43\, Y^2 \Big)\Big( 1 - 3.87\, Z^2 \Big) \label{eq_u_noise} \\[0.5cm]
p^{\text{symb\_noise}} &= \mathrm{Re}\, \Big( 143.42 - 28.68\, X \Big) \label{eq_p_noise}
\end{align}
\end{center}

\vspace{4mm}

Both Equations~\eqref{eq_u_noise} and~\eqref{eq_p_noise} remain very close to their noise-free counterparts (Equations~\eqref{eq_u} and~\eqref{eq_p}, respectively), demonstrating strong robustness to perturbations in the data. This is further confirmed by the corresponding MSE, MAE, and NMAE values on both the training and testing datasets, which exhibit only minor deviations from the deterministic case (cf. Table~\ref{table_equations}).

In addition, we verified that similar levels of accuracy and stability are obtained when expanding the SR operator set to include trigonometric and exponential functions (cf. Table~\ref{table_sr_parameters}); even within these richer search spaces, the SR procedure consistently converged to polynomial-type expressions, indicating that the discovered relationships are not an artefact of a restricted operator set, but a genuine reflection of the underlying flow physics.

\section[Symbolic Regression Enhanced with Answer Set Programming]{Symbolic Regression Enhanced with Answer Set Programming}
\label{section_sr_asp}

As already made clear, SR has gained prominence as a robust methodology for discovering mathematical relationships within complex datasets, without presupposing a specific model structure. While SR excels in its generative capabilities, it often operates within a {\em purely data-driven} paradigm, potentially overlooking intricate domain-specific constraints and logical relationships inherent to physical systems. This limitation can lead to the selection of models that, despite their statistical accuracy, {\em may violate} fundamental physical laws or lack interpretability within the context of established scientific theories. To address these challenges, we propose an integration of SR with Answer Set Programming (ASP) --- a form of {\em declarative} programming oriented towards difficult combinatorial search problems \cite{lifschitz19,brewka11,gebser12}. This innovative {\em hybrid} approach seamlessly merges data-driven and symbolic Artificial Intelligence into a unified framework, combining the generative power of SR with the declarative reasoning strengths of ASP, which not only enhances model selection, but also ensures adherence to physical principles (refer to Figure~\ref{fig_sr_asp} of Introduction). It is important to highlight that, although both ASP and SR have been extensively studied independently, our proposal represents, to the best of our knowledge, the first attempt to integrate these two powerful frameworks.

\subsection{Answer Set Programming}
\label{subsection_asp}

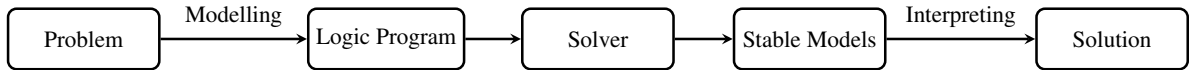
\begin{figure}[h!]
\centering
\scalebox{0.8}{
\begin{tikzpicture}[very thick]
\node (1) [draw, shape=rectangle, rounded corners, align=center, minimum width=2.5cm, minimum height = 1cm] at (-8.5,0) {Problem};
\node (2) [draw, shape=rectangle, rounded corners, align=center, minimum width=2.5cm, minimum height = 1cm] at (-3.5,0) {Logic Program};
\node (3) [draw, shape=rectangle, rounded corners, align=center, minimum width=2.5cm, minimum height = 1cm] at (0,0) {Solver};
\node (4) [draw, shape=rectangle, rounded corners, align=center, minimum width=2.5cm, minimum height = 1cm] at (3.5,0) {Stable Models};
\node (5) [draw, shape=rectangle, rounded corners, align=center, minimum width=2.5cm, minimum height = 1cm] at (8.5,0) {Solution};

\draw[->, >=stealth] (1) -- (2) node[midway, above=0.5mm]{Modelling};
\draw[->, >=stealth] (2) -- (3);
\draw[->, >=stealth] (3) -- (4);
\draw[->, >=stealth] (4) -- (5) node[midway, above=0.5mm]{Interpreting};
\end{tikzpicture}}
\caption{The solving process of ASP \cite{gebser12}.}
\label{fig_asp_solving}
\end{figure}

ASP constitutes a {\em knowledge-representation} paradigm that allows the specification of complex problems and constraints in a high-level, declarative language. Unlike imperative approaches, ASP focuses on describing {\em what} the solution should satisfy rather than {\em how} to discover it \cite{lifschitz19,brewka11,gebser12}. As visualized in Figure~\ref{fig_asp_solving}, ASP operates by defining a logic program, namely, {\em rules} and {\em facts} that describe the problem domain, allowing an ASP solver to compute {\em stable models}, also known as {\em answer sets} --- i.e., consistent sets of literals (atomic formulae or their negation) that satisfy all given constraints. This capability makes ASP particularly suitable for tasks that require intricate constraint satisfaction, logical inference, and combinatorial optimization \cite{erdem16,falkner18}.

\vspace{3mm}

\subsection{Methodology}
\label{subsection_methodology}

The integration of ASP with SR involves a systematic work-flow that leverages the strengths of both methodologies. Initially, SR tools like \texttt{PySR} are employed to explore the space of possible mathematical models that fit the fluid-mechanics data. This process yields a diverse set of candidate equations, each characterized by attributes such as complexity, loss (an error metric), and functional form. However, the raw output from SR often includes models that may not comply with domain-specific constraints essential for physical validity.

To imbue the SR-generated models with domain-specific intelligence, the next step involves preprocessing and feature extraction. This can be accomplished using scripting languages like Python, where the symbolic forms of the candidate equations are parsed to extract relevant features. Features of interest may include the presence of specific terms (e.g., $X^4$), functional properties (such as monotonicity or asymptotic behaviour), and dimensional attributes ensuring dimensional consistency. These extracted features are then represented as facts in an ASP-compatible format, effectively translating the symbolic information into a form that ASP can process.

Subsequently, an ASP program can be developed to encapsulate the desired constraints and domain-specific knowledge. These constraints can range from simple exclusions or inclusion of certain terms to more complex rules enforcing physical laws such as conservation of mass and momentum. For instance, the ASP program can include rules that prevent the selection of equations containing non-physical terms (e.g., $X^4$), enforce the selection of equations containing essential parameters like the Reynolds number ($\mathrm{Re}$), or ensure that selected models adhere to dimensional homogeneity. Additionally, ASP's optimization capabilities can be harnessed to balance multiple objectives, such as minimizing both model complexity and prediction error while ensuring compliance with physical constraints.

An ASP solver, such as \textsc{Clasp}, is then employed to process (the propositional form of) the ASP program, alongside the preprocessed facts derived from the SR output. \textsc{Clasp} is a widely used, off-the-shelf ASP solver that incorporates state-of-the-art techniques from Boolean constraint solving \cite{Gebser07}. The solver evaluates the rules and constraints, identifying {\em stable models} that represent valid selections of equations adhering to all specified criteria. These stable models effectively filter out non-compliant equations, retaining only those that are both data-accurate and physically meaningful. The (physically) validated models can then be utilized for further analysis, simulations, or as components of more comprehensive fluid-mechanics theories.

\subsection{A Concrete Example}
\label{subsection_asp_example}

To demonstrate the practical integration of the proposed SR/ASP hybrid approach, we present a concrete example based on the equations derived by the SR model tailored to the axial velocity $u$ of the flow (see Section~\ref{section_results}). Recall that these equations are all part of the {\em Pareto front} of the SR procedure, meaning that each one retains an optimal balance of accuracy and simplicity. The presented example illustrates how SR-generated symbolic expressions can be effectively filtered and validated using ASP to ensure their physical plausibility and adherence to domain-specific constraints. As will be shown, the methodology is highly versatile, allowing for the incorporation of numerous constraints or adaptation to other domains.

First, Table~\ref{table_sr_equations} reports all the ($14$) equations of the Pareto front, identified by the alluded SR model. Each equation varies in complexity and loss (MSE), representing different levels of fit and intricacy in capturing the underlying physical phenomena. Note that
the equation corresponding to $\textbf{ID 9}$ is Equation~\ref{eq_u} of Section~\ref{section_results}.

\begin{table}[t]
\centering\scriptsize
\renewcommand{\arraystretch}{1.6}
\begin{tabular}{|M{0.3cm}|M{1.5cm}|M{2cm}|M{8cm}|}
\hline
\textbf{ID} & \textbf{Complexity} & \textbf{Loss} & \textbf{SR Equation for the Axial Velocity} $\textbf{u}$ \\ \hline\hline
$0$ & $1$ & $37396.16$ & $X$ \\ \hline
$1$ & $2$ & $31757.25$ & $X^3$ \\ \hline
$2$ & $3$ & $13316.42$ & $0.91 \, \mathrm{Re}$ \\ \hline
$3$ & $5$ & $13316.14$ & $0.91 \, \mathrm{Re} - 0.63$ \\ \hline
$4$ & $6$ & $10670.45$ & $\mathrm{Re} - 427.88\, Y^2$ \\ \hline
$5$ & $8$ & $9246.47$ & $\mathrm{Re} - 849.34\, Y^2 + 76.16$ \\ \hline
$6$ & $10$ & $7431.94$ & $1.41\, \mathrm{Re} - 5.62\, \mathrm{Re}\, Z^2$ \\ \hline
$7$ & $13$ & $7417.14$ & $1.41\, \mathrm{Re} - 5.62\, \mathrm{Re} \, Z^2  - 1.41 \, Y^2$ \\ \hline
$8$ & $15$ & $3620.02$ & $(1.70 - 7.35\, Y^2)\, (\mathrm{Re} - 400.53 \, Z^2) + 11.5$ \\ \hline
$\textbf{9}$ & $\textbf{17}$ & \textbf{44.98} & $\mathbf{Re} \left( \mathbf{2.18} - \mathbf{8.46}\, \mathbf{Y^2} \right) \left( \mathbf{1} - \mathbf{3.89}\, \mathbf{Z^2} \right)$ \\ \hline
$10$ & $20$ & $25.31$ & $(2.15 -8.45 \, Y^2 + Z^2)\, (\mathrm{Re} -3.98 \, \mathrm{Re} \, Z^2) + 0.35$ \\ \hline
$11$ & $22$ & $25.07$ & $(2.15 - 8.45 \, Y^2 + Z^2)\, (\mathrm{Re}  -3.98 \, \mathrm{Re} \, Z^2 + 0.53) - 0.41$ \\ \hline
$12$ & $24$ & $25.02$ & $(2.15 - 8.45 \, Y^2 + Z^2) \, (\mathrm{Re} -3.98 \, \mathrm{Re} \, Z^2 + Y^2) - Y^3$ \\ \hline
$13$ & $25$ & $22.96$ & $(2.15 - 8.45 \, Y^2 + Z^2) \, (\mathrm{Re} -3.98 \, \mathrm{Re} \, Z^2 + 8.45 \, Y^2) - Y$ \\ \hline
\end{tabular}
\caption{The equations of the Pareto front, identified by the SR model tailored to the axial velocity $u$ of the flow. Observe that the complexity of equations remains up to $25$, as constrained by the corresponding limitation outlined in Table~\ref{table_sr_parameters} of Section~\ref{section_sr_model}.}
\label{table_sr_equations}
\end{table}

While the symbolic expressions of Table~\ref{table_sr_equations} achieve an optimal balance between accuracy and simplicity, they do {\em not necessarily} adhere to physical laws and/or domain-specific constraints. To ensure such an adherence, ASP will be employed as a filtering mechanism that enforces the desired constraints. To that end, the SR output is, first, translated into ASP-compatible facts. Additionally, the presence of specific terms that may violate physical constraints, such as non-physical exponents or the inclusion of essential parameters like the Reynolds number ($\mathrm{Re}$) can be identified. Such an ASP-encoding is, indicatively, presented subsequently for the equations corresponding to $\textbf{ID 7}$ and $\textbf{ID 9}$ of Table~\ref{table_sr_equations}. Note that, as ASP cannot process real numbers, the loss values have been rounded to the nearest integers for compatibility. For a comprehensive discussion of ASP's syntax and semantics, the interested reader is referred to the Potassco guide \cite{potassco17}.

\vspace{3mm}

{\footnotesize\centering\noindent
\begin{tcolorbox}[colback=white, width=\columnwidth, colframe=black, arc=3mm, boxrule=1pt, left=2mm]
\begin{verbatim}
% SR equations as facts
eq(7, 13, 7417, "1.41*Re-5.62*Re*Z**2-1.41*Y**2").
contains_re(7).
contains_y2(7).
contains_z2(7).

eq(9, 17, 45, "Re*(2.18-8.46*Y**2)*(1-3.89*Z**2)").
contains_re(9).
contains_y2(9).
contains_z2(9).
\end{verbatim}
\end{tcolorbox}}

\vspace{2mm}

In the above code snippet, the predicate \texttt{eq/4} represents an SR equation in the form \mbox{\texttt{eq(ID, Complexity, Loss, "Expression")}}, where the four arguments encapsulate essential information about the equation. The first argument, \texttt{ID}, uniquely identifies the equation within the dataset, allowing for efficient referencing and filtering. The second argument, \texttt{Complexity}, denotes the complexity of the equation. The third argument, \texttt{Loss}, specifies the MSE of the equation when evaluated against the dataset. Finally, the fourth argument, \texttt{"Expression"}, is a string representation of the symbolic equation itself. An indicative fact that complements an equation is expressed via the predicate \texttt{contains\_re(7)}, which states that the equation corresponding to $\textbf{ID 7}$ includes the Reynolds number (\(\mathrm{Re}\)).

Thereafter, we can impose constraints such as maximum allowable complexity and loss, exclusion of non-physical terms (e.g., $X^3$, $Y^3$, $X^4$), and the requirement that selected models incorporate essential physical parameters like the Reynolds number ($\mathrm{Re}$). Such requirements can be represented in an ASP-format as follows:

\vspace{4mm}

{\footnotesize\centering\noindent
\begin{tcolorbox}[colback=white, width=\columnwidth, colframe=black, arc=3mm, boxrule=1pt, left=1.5mm]
\begin{verbatim}
% Define maximum allowed complexity and loss for eligible equations
#const max_complexity = 20.
#const max_loss = 100.

% Choice rule: Any subset of equations can be chosen, subject to constraints
{ chosen(EqID) : eq(EqID, _, _, _) }.

% Eligibility criteria: An equation is eligible if its complexity and loss 
% are within the specified limits
eligible(EqID) :- eq(EqID, Complexity, Loss, _), 
                  Complexity <= max_complexity, 
                  Loss <= max_loss.

% Constraint: Only select eligible equations
:- chosen(EqID), not eligible(EqID).

% Constraint: Exclude equations containing certain terms
:- chosen(EqID), contains_x3(EqID).
:- chosen(EqID), contains_y3(EqID).
:- chosen(EqID), contains_x4(EqID).

% Constraint: Ensure at least one chosen equation contains Re
:- not { chosen(EqID) : contains_re(EqID) }.
\end{verbatim}
\end{tcolorbox}}

\vspace{2mm}

Upon formulating the ASP program, the \textsc{Clasp} ASP solver is employed to process the program, and identify stable models (answer sets) that satisfy all the imposed constraints, evaluating each candidate equation against the eligibility criteria and exclusion constraints. After executing the ASP solver, we derive the following selected/filtered equations in approximately 0.1 seconds --- notably, the first equation corresponds to Equation~\ref{eq_u} of Section~\ref{section_results}.

\vspace{4mm}

{\footnotesize\centering\noindent
\begin{tcolorbox}[colback=white, width=\columnwidth, colframe=black, arc=3mm, boxrule=1pt, left=1.5mm]
\begin{verbatim}
SELECTED EQUATIONS
------------------

ID 9: Re*(2.18-8.46*Y**2)*(1-3.89*Z**2)
Complexity: 17, Loss: 45

ID 10: (2.15-8.45*Y**2+Z**2)*(Re-3.98*Re*Z**2)+0.35
Complexity: 20, Loss: 25
\end{verbatim}
\end{tcolorbox}}

\vspace{2mm}

Evidently, the expressions appearing in the list of ``\texttt{SELECTED EQUATIONS}'' strike an ideal trade-off between accuracy and simplicity (as they are part of the Pareto front of the SR procedure) and, at the same time, they are physically valid (as they adhere to the imposed domain-specific constraints of ASP).

The computation time required by the \textsc{Clasp} ASP solver to generate stable models in our representative example is indeed negligible --- approximately 0.1 seconds, which is significantly less than the computation time (around 15 minutes) required by the SR model to derive the equations of the Pareto front (see Section~\ref{section_results}). Nevertheless, in more complex fluid-flow problems and geometries, the sheer volume of potential SR equations and the nature of domain-specific constraints can significantly enlarge the search space for stable models and, by extension, increase computational overhead. To mitigate these concerns, one can exploit incremental solving strategies (where constraints are added iteratively), utilize domain-decomposition techniques to tackle sub-problems in parallel, or take advantage of advanced ASP features, such as heuristic-guided optimization, which can substantially narrow the search space \cite{gebser12}. These practices can allow the introduced framework to remain viable in larger, more intricate fluid-mechanics scenarios, while keeping computation times in check.

\subsection{More Expressive Physical Constraints}

The constraints discussed so far are mostly ``syntactic'', but ASP's expressive language can also encode genuinely physical constraints. In this subsection, we present a representative collection of such constraints.

\vspace{5mm}

\noindent{\bf No-Slip Boundary Condition}

\vspace{1mm}

\begin{center}
\begin{verbatim}
:- chosen(EqID), wall(P), val(u, EqID, P, V), |V| > eps_u.
\end{verbatim}
\end{center}

\vspace{1mm}

In this constraint, the predicate \texttt{wall(P)} marks each point \texttt{P} that lies on a duct wall (these points are defined in advance). The predicate \texttt{val(u, EqID, P, V)} provides the numerically evaluated value \texttt{V} of the candidate velocity equation \texttt{EqID} at point \texttt{P}. The comparison \texttt{|V| > eps\_u} checks whether the magnitude of this value exceeds a small predefined tolerance \texttt{eps\_u}, used to represent ``effectively zero''. If all these conditions hold simultaneously, the candidate equation violates the no-slip boundary condition. Consequently, ASP eliminates any equation that produces a non-zero wall velocity, thereby ensuring that admissible symbolic expressions satisfy the requirement that the axial velocity vanishes at all wall locations.

\vspace{5mm}

\noindent{\bf Centerline Symmetry}

\vspace{1mm}

\begin{center}
\begin{verbatim}
:- chosen(EqID), center_y(P), dudy(EqID, P, G), |G| > eps_g.
:- chosen(EqID), center_z(P), dudz(EqID, P, G), |G| > eps_g.
\end{verbatim}
\end{center}

\vspace{1mm}

In a rectangular duct, the velocity field is symmetric with respect to the mid-planes \(Y = 0\) and \(Z = 0\). This symmetry implies that the transverse gradients of the axial velocity must vanish at these locations, that is,
\[
\frac{\partial u}{\partial Y} = 0 \quad \text{at} \quad Y = 0
\qquad \text{and} \qquad
\frac{\partial u}{\partial Z} = 0 \quad \text{at} \quad Z = 0.
\]
Accordingly, in the above constraint, the predicate \texttt{center\_y(P)} (resp., \texttt{center\_z(P)}) identifies every point \texttt{P} located on the symmetry plane \(Y=0\) (resp., \(Z=0\)). The predicates \mbox{\texttt{dudy(EqID, P, G)}} and \texttt{dudz(EqID, P, G)} represent the evaluated transverse gradients \(\partial u/\partial Y\) and \(\partial u/\partial Z\) of the candidate velocity equation \texttt{EqID} at point \texttt{P}. The comparison \texttt{|G| > eps\_g} checks whether the magnitude of these gradients exceeds a small tolerance \texttt{eps\_g}. Whenever this occurs, the constraint is triggered and the candidate equation is eliminated. In this way, the ASP module filters out symbolic expressions that violate the geometric symmetry of the flow.

\vspace{5mm}

\noindent{\bf Symmetry and Laminar Scaling}

\vspace{1mm}

\begin{center}
\begin{verbatim}
:- chosen(EqID), has_term(EqID, y_pow(K)), K #mod 2 == 1.
:- chosen(EqID), has_term(EqID, z_pow(K)), K #mod 2 == 1.
:- chosen(EqID), has_term(EqID, re_pow(K)), K != 1.
\end{verbatim}
\end{center}

\vspace{1mm}

The geometry of the rectangular duct is symmetric with respect to the transverse coordinates \(Y\) and \(Z\), and the corresponding laminar velocity field inherits this symmetry. As a consequence, physically admissible expressions for the axial velocity must depend only on even powers of \(Y\) and \(Z\). In addition, in the considered laminar regime, the axial velocity scales linearly with the Reynolds number \(\mathrm{Re}\), and thus non-linear powers of \(\mathrm{Re}\) are physically inconsistent. In the ASP rules above, the predicate \texttt{has\_term(EqID, y\_pow(K))} (resp., \texttt{z\_pow(K)}) indicates that the symbolic expression associated with the equation \texttt{EqID} contains a term proportional to \(Y^K\) (resp., \(Z^K\)). The condition \mbox{\texttt{K \#mod 2 == 1}} (modulo operation) identifies odd exponents, which violate geometric symmetry. Similarly, the predicate \texttt{has\_term(EqID, re\_pow(K))} detects terms involving powers of \(\mathrm{Re}\), and the condition \mbox{\texttt{K != 1}} filters out non-linear dependencies. Whenever any of these conditions hold, the corresponding constraint is triggered and the candidate equation is rejected. This ensures that the ASP module retains only symbolic expressions that respect the symmetry of the flow and the appropriate laminar scaling with the Reynolds number.

\subsection{Final Remarks}

We conclude our discussion by noting that the proposed approach is general and can, in principle, recover different analytical forms when applied to datasets reflecting other physical scenarios. By adjusting the SR search space and tailoring the ASP constraints to the corresponding boundary and physical requirements, the framework can be straightforwardly adapted to alternative geometries, flow regimes, or governing equations, without altering its core methodology. We also note that the SR/ASP hybrid framework can, to a certain extent, be paralleled with {\em Physics-Informed Neural Networks} (PINNs) \cite{raissi19,cai21}. These specialized Artificial Neural Networks embed physical laws directly into their learning process, producing models that are inherently consistent with the underlying physics of the application. Analogously, the SR/ASP hybrid approach capitalizes on domain knowledge to refine and validate the SR equations, thus achieving greater alignment with the physical principles governing a problem.

\section{Conclusions}

Recognizing that a thorough understanding of the underlying flow physics is as crucial as accurate prediction in fluid mechanics, we present in this article a Symbolic-Regression (SR) approach for modelling fluid dynamics; particularly, a laminar flow within a three-dimensional (3D) channel. The study focused on deriving symbolic equations that describe the velocity and pressure fields, leveraging simulation data generated with the Finite Volume Method (FVM) to achieve an interpretable yet accurate representation of the flow dynamics.

The channel geometry consisted of a rectangular cross-section, with flow characterized by a parabolic velocity profile and a linear pressure drop along the channel length, typical of fully developed laminar flow. Using the \texttt{PySR} library and without prescribing a specific functional form in advance, symbolic equations for the axial velocity ($u$) and pressure ($p$) fields were derived and evaluated. The velocity equation demonstrated a quadratic dependence on the lateral ($Y$) and vertical ($Z$) spatial variables, which define the channel's cross-sectional plane, consistent with the physical principles of laminar flow, while the pressure equation captured the expected linear drop along the longitudinal direction ($X$).

Quantitative evaluation of the derived models highlighted their robustness and accuracy. The symbolic equation for velocity, Equation~\eqref{eq_u}, achieved a Normalized Mean Absolute Error (NMAE) of less than $0.01\%$ on both the training and testing datasets, showcasing its ability to generalize effectively across unseen data. Similarly, the pressure model, Equation~\eqref{eq_p}, yielded a NMAE on the order of \( 10^{-8} \% \), underscoring its precision in capturing the channel's pressure gradient. These results not only validate the fidelity of the SR-derived models, but also affirm their potential for practical applications. The derived symbolic equations were further compared against well-established analytical solutions from the literature. The remarkable agreement between the SR-generated equations and the analytical velocity profile, as well as the linear pressure distribution in rectangular ducts, underscores the capability of SR to distil physically meaningful relationships directly from simulation raw data.

Although, SR demonstrates remarkable generative capabilities, its purely data-driven approach can overlook complex domain-specific constraints and logical relationships inherent to physical systems. This shortcoming may result in the selection of symbolic expressions that, while statistically accurate, fail to comply with fundamental physical laws or other critical domain-specific principles. To overcome these limitations, we proposed a novel integration of SR with the knowledge-representation framework of Answer Set Programming (ASP). By merging data-driven and symbolic Artificial Intelligence, the introduced hybrid SR/ASP framework ensures that the symbolic expressions generated by the SR models are not only statistically sound, but also adhere to physical plausibility and domain-specific rules, symbolically encoded into ASP.

In future studies, we aim to extend our SR approach to encompass more intricate fluid-flow scenarios and geometries. Additionally, we plan to integrate Chebyshev polynomials as foundational elements (operators) within the SR algorithm. This choice is motivated by the established utility of Chebyshev polynomials in spectral methods --- powerful numerical techniques for solving partial differential equations \cite{canuto88,hussaini87}. Their proven effectiveness in accurately modelling fluid flows makes them an excellent candidate for enhancing the robustness and precision of our SR framework. As far as the ASP module is concerned, an appealing avenue for future research involves the development of methods that automatically derive constraints from extensive domain knowledge bases, thereby minimizing the need for manual constraint encoding. In this context, the contribution of Large Language Models would be invaluable, given their ability to process and interpret natural-language descriptions (of domain-specific information), and translate them into formal representations (relevant rules and constraints) compatible with ASP \cite{leong2023,endres2024}.

\section*{Acknowledgments} 

This research was implemented in the framework of the Action ``Flagship actions in interdisciplinary scientific fields with a special focus on the productive fabric'', which is implemented through the National Recovery and Resilience Fund Greece 2.0 and funded by the European Union--NextGenerationEU (Project ID: TAEDR-0535983).

The authors gratefully acknowledges the anonymous reviewers for their insightful and constructive comments on an earlier version of this article.

\vspace{1cm}

\noindent {\bf Author Contributions:} All authors contributed equally to this work.

\vspace{3mm}

\noindent {\bf Competing Interests:} The authors declare no competing interests.

\vspace{3mm}

\noindent {\bf Data Availability:} The datasets generated during the current study are available on reasonable request.

\currentpdfbookmark{References}{References}\bibliographystyle{plain}
\bibliography{references}

@book{kronberger24,
  author = "Gabriel Kronberger and Bogdan Burlacu and Michael Kommenda and Stephan M. Winkler and Michael Affenzeller",
  title = "Symbolic Regression",
  publisher = "CRC Press",
  address = "",
  year = "2024"
}

@book{petrowski17,
  author = "Alain P{\'e}trowski and Sana Ben-Hamida",
  title = "Evolutionary Algorithms",
  publisher = "Wiley",
  address = "",
  year = "2017"
}

@article{schmidt09,
  author = "Michael Schmidt and Hod Lipson",
  title = "Distilling free-form natural laws from experimental data",
  journal = "Science",
  volume = "324",
  pages = "81--85",
  year = "2009"
}

@article{udrescu20,
  author = "Silviu-Marian Udrescu and Max Tegmark",
  title = "{AI} {F}eynman: {A} physics-inspired method for symbolic regression",
  journal = "Science Advances",
  volume = "6",
  pages = "",
  year = "2020"
}

@article{makke24,
  author = "Nour Makke and Sanjay Chawla",
  title = "Interpretable scientific discovery with symbolic regression: {A} review",
  journal = "Artificial Intelligence Review",
  volume = "57",
  number = "2",
  pages = "",
  year = "2024"
}

@article{angelis23,
  author = "Dimitrios Angelis and Filippos Sofos and Theodoros E. Karakasidis",
  title = "Artificial {I}ntelligence in physical sciences: {S}ymbolic regression trends
and perspectives",
  journal = "Archives of Computational Methods in Engineering",
  volume = "30",
  number = "",
  pages = "3845--3865",
  year = "2023"
}

@article{kim21,
  author = "Samuel Kim and Peter Y. Lu and Srijon Mukherjee and Michael Gilbert and Li Jing and Vladimir \u{C}eperi{\'c} and Marin Solja\u{c}i{\'c}",
  title = "Integration of neural network-based symbolic regression in deep learning for scientific discovery",
  journal = "IEEE Transactions on Neural Networks and Learning Systems",
  volume = "32",
  pages = "",
  year = "2021"
}

@article{cranmer23,
  author = "Miles Cranmer",
  title = "Interpretable machine learning for science with {P}y{SR} and {S}ymbolic{R}egression.jl",
  journal = "arXiv:2305.01582",
  volume = "",
  pages = "",
  year = "2023"
}

@book{wang11,
  author = "Lihui Wang and Amos H. C. Ng and Kalyanmoy Deb",
  title = "Multi-Objective Evolutionary Optimisation for Product Design and Manufacturing",
  publisher = "Springer",
  address = "London",
  year = "2011"
}

@article{roland19,
  author = "Wolfgang Roland and Christian Marschik and Michael Krieger and Bernhard L{\"o}w-Baselli and J{\"u}rgen Miethlinger",
  title = "Symbolic regression models for predicting viscous dissipation of three-dimensional non-{N}ewtonian flows in single-screw extruders",
  journal = "Journal of Non-Newtonian Fluid Mechanics",
  volume = "268",
  pages = "12--29",
  year = "2019"
}

@article{wu23,
  author = "Chenyu Wu and Yufei Zhang",
  title = "Enhancing the shear-stress-transport turbulence model with symbolic regression: {A} generalizable and interpretable data-driven approach",
  journal = "Physical Review Fluids",
  volume = "8",
  number = "084604",
  pages = "",
  year = "2023"
}

@article{angelis24,
  author = "Dimitrios Angelis and Filippos Sofos and Theodoros E. Karakasidis",
  title = "Reassessing the transport properties of fluids: {A} symbolic regression approach",
  journal = "Physical Review E",
  volume = "109",
  number = "015105",
  pages = "",
  year = "2024"
}

@article{hasadi19,
  author = "Yousef M. F. El Hasadi and Johan T. Padding",
  title = "Solving fluid flow problems using semi-supervised symbolic regression on sparse data",
  journal = "AIP Advances",
  volume = "9",
  number = "115218",
  pages = "",
  year = "2019"
}

@article{milosevic22,
  author = "Marko Milo{\v{s}}evi{\'c} and Dejan Brki{\'c} and Pavel Praks and Dragan Litri{\v{c}}in and Zoran Staji{\'c}",
  title = "Hydraulic losses in systems of conduits with flow from laminar to fully turbulent: {A} new symbolic regression formulation",
  journal = "Axioms",
  volume = "11",
  number = "5",
  pages = "",
  year = "2022"
}

@article{sofos22,
  author = "Filippos Sofos and Avraam Charakopoulos and Konstantinos Papastamatiou and Theodoros E. Karakasidis",
  title = "A combined clustering/symbolic regression framework for fluid property prediction",
  journal = "Physics of Fluids",
  volume = "34",
  number = "062004",
  pages = "",
  year = "2022"
}

@article{praks18,
  author = "Pavel Praks and Dejan Brki{\'c}",
  title = "Symbolic regression-based genetic approximations of the {C}olebrook equation for flow friction",
  journal = "Water",
  volume = "10",
  number = "9",
  pages = "",
  year = "2018"
}

@article{reinbold21,
  author = "Patrick A. K. Reinbold and Logan M. Kageorge and Michael F. Schatz and Roman O. Grigoriev",
  title = "Robust learning from noisy, incomplete, highdimensional experimental data via physically constrained symbolic regression",
  journal = "Nature Communications",
  volume = "12",
  number = "3219",
  pages = "",
  year = "2021"
}

@article{alam22,
  author = "Todd M. Alam and Joshua P. Allers and Calen J. Leverant and Jacob A. Harvey",
  title = "Symbolic regression development of empirical equations for diffusion in {L}ennard-{J}ones fluids",
  journal = "The Journal of Chemical Physics",
  volume = "157",
  number = "014503",
  pages = "",
  year = "2022"
}

@article{chakrabarty21,
  author = "A. Chakrabarty and Sergey N. Yakovenko",
  title = "Data-driven turbulence modelling using symbolic regression",
  journal = "Journal of Physics: Conference Series",
  volume = "2099",
  number = "012020",
  pages = "",
  year = "2021"
}

@article{raissi19,
  author = "Maziar Raissi and Paris Perdikaris and George Em. Karniadakis",
  title = "Physics-informed neural networks: {A} deep learning framework for solving forward and inverse problems involving nonlinear partial differential equations",
  journal = "Journal of Computational Physics",
  volume = "378",
  pages = "686--707",
  year = "2019"
}

@article{cai21,
  author = "Shengze Cai and Zhiping Mao and Zhicheng Wang and Minglang Yin and George Em. Karniadakis",
  title = "Physics-informed neural networks ({PINN}s) for fluid mechanics: {A} review",
  journal = "Acta Mechanica Sinica",
  volume = "37",
  pages = "1727--1738",
  year = "2021"
}

@book{kakac87,
  title = "Handbook of Single-Phase Convective Heat Transfer",
  author = "Sadik Kaka\c{c} and Ramesh K. Shah and Win Aung",
  year = "1987",
  edition   = "",
  publisher = "Wiley-Interscience"
}

@book{pritchard11,
  title = "Fox and McDonald's Introduction to Fluid Mechanics",
  author = "Philip J. Pritchard and John C. Leylegian",
  year = "2011",
  edition   = "8th",
  publisher = "John Wiley \& Sons"
}

@book{white11,
  title = "Fluid Mechanics",
  author = "Frank M. White",
  year = "2011",
  edition   = "7th",
  publisher = "McGraw Hill"
}

@book{moukalled16,
  title = "The Finite Volume Method in Computational Fluid Dynamics: An Advanced Introduction with OpenFOAM\textsuperscript{\textregistered} and Matlab",
  author = "Fadl Moukalled and Luca Mangani and Marwan Darwish",
  year = "2016",
  publisher = "Springer"
}

@article{tzirtzilakis12004biomagnetic,
  title={Biomagnetic fluid flow in a 3{D} rectangular duct},
  author={E. E. Tzirtzilakis and V. D. Sakalis and N. G. Kafoussias and P. M. Hatzikonstantinou},
  journal={International Journal for Numerical Methods in Fluids},
  volume={44},
  number={12},
  pages={1279--1298},
  year={2004},
  publisher={Wiley Online Library}
}

@article{aravanis24b,
  author = "Theofanis Aravanis and Grigorios Chrimatopoulos and Michalis Xenos and Efstratios E. Tzirtzilakis",
  title = "Forecasting two-dimensional channel flow using {M}achine {L}earning",
  journal = "Physics of Fluids",
  volume = "36",
  number    = {103617},
  pages = "",
  year = "2024",
  publisher = {American Institute of Physics}
}

@article{hussaini87,
  author    = {M Y Hussaini and Thomas A. Zang},
  title     = {Spectral methods in fluid dynamics},
  journal   = {Annual Review of Fluid Mechanics},
  volume    = {19},
  number    = {},
  pages     = {339--367},
  year      = {1987}
}

@book{canuto88,
  author    = {Claudio Canuto and M. Yousuff Hussaini and Alfio Quarteroni and Thomas A. Zang},
  title     = {Spectral Methods in Fluid Dynamics},
  edition   = {},
  year      = {1988},
  publisher = {Springer},
  address   = {}
}

@book{gebser12,
  author = "Martin Gebser and Ronald Kaminski and Benjamin Kaufmann and Torsten Schaub",
  title = "Answer Set Solving in Practice",
  series   = "Synthesis Lectures on Artificial Intelligence and Machine Learning",
  publisher = "Morgan \& Claypool Publishers",
  address = "",
  number = "",
  year = "2012"
}

@book{lifschitz19,
  author = "Vladimir Lifschitz",
  title = "Answer Set Programming",
  publisher = "Springer",
  address = "",
  year = "2019"
}

@article{brewka11,
  author = "Gerhard Brewka and Thomas  Eiter and Miros\l{}aw Truszczy{\'n}ski",
  title = "Answer {S}et {P}rogramming at a glance",
  journal = "Communications of the ACM",
  volume = "54",
  number = "12",
  pages = "93--103",
  year = "2011"
}

@article{erdem16,
  author = "Esra Erdem and Michael Gelfond and Nicola Leone",
  title = "Applications of {A}nswer {S}et {P}rogramming",
  journal = "AI Magazine",
  volume = "37",
  number = "3",
  pages = "53--68",
  year = "2016"
}

@article{falkner18,
  author = "Andreas Falkner and Gerhard Friedrich and Konstantin Schekotihin and Richard Taupe and Erich C. Teppan",
  title = "Industrial applications of {A}nswer {S}et {P}rogramming",
  journal = "KI -- K{\"u}nstliche Intelligenz",
  volume = "32",
  number = "",
  pages = "165--176",
  year = "2018"
}

@inproceedings{Gebser07,
  author = "Martin Gebser and Benjamin Kaufmann and Andr{\'e} Neumann and Torsten Schaub",
  title = "Clasp: A conflict-driven answer set solver",
  booktitle = "Logic Programming and Nonmonotonic Reasoning (LPNMR 2007), Lecture Notes in Computer Science",
  editor = "Chitta Baral and Gerhard Brewka and John Schlipf",
  pages = "260--265",
  publisher = "Springer",
  address = "",
  month = "",
  year = "2007"
}

@manual{potassco17,
  author = "Martin Gebser and Roland Kaminski and Benjamin Kaufmann and Marius Lindauer and Max Ostrowski and Javier Romero and Torsten Schaub and Sven Thiele",
  title = "Potassco User Guide",
  organization = "Potassco",
  edition  = "2nd",
  address = "University of Potsdam",
  year = "2017"
}

@inproceedings{leong2023,
  author    = {Iat Tou Leong and Raul Barbosa},
  title     = {Translating natural language requirements to formal specifications: {A} study on {GPT} and symbolic {NLP}},
  booktitle = {2023 53rd Annual IEEE/IFIP International Conference on Dependable Systems and Networks Workshops (DSN-W)},
  year      = {2023},
  publisher = {IEEE},
  pages     = {259--262}
}

@article{endres2024,
  author    = {Madeline Endres and Sarah Fakhoury and Saikat Chakraborty and Shuvendu K. Lahiri},
  title     = {Can large language models transform natural language intent into formal method postconditions?},
  journal   = {Proceedings of the ACM on Software Engineering},
  volume    = {1},
  article   = {84},
  year      = {2024},
  pages     = {24},
  publisher = {ACM}
}

@inproceedings{platzer24,
  author = "Andr{\'e} Platzer",
  title = "Intersymbolic {AI}",
  booktitle = "Leveraging Applications of Formal Methods, Verification and Validation. Software Engineering Methodologies, ISoLA 2024",
  editor = "Tiziana Margaria and Bernhard Steffen",
  pages = "1070--1080",
  publisher = "Springer",
  address = "",
  year = "2024"
}

@book{mazumder2015numerical,
  title={Numerical Methods for Partial Differential Equations: Finite Difference and Finite Volume Methods},
  author={Sandip Mazumder},
  year={2016},
  publisher={Academic Press}
}

@book{munson2012fundamentals,
  title={Fundamentals of Fluid Mechanics},
  author={Bruce R. Munson and Alric P. Rothmayer and Theodore H. Okiishi and Wade W. Huebsch},
  year={2012},
  edition = {7th},
  publisher={Wiley}
}

@article{chrimatopoulos2023,
  title={Magnetohydrodynamic and ferrohydrodynamic fluid flow using the finite volume method},
  author={Grigorios Chrimatopoulos and Efstratios E. Tzirtzilakis and Michalis A. Xenos},
  journal={Fluids},
  volume={9},
  number={1},
  year={2024},
  publisher={MDPI}
}

@article{katsoudas2025,
  title={A finite volume and {L}evenberg-{M}arquardt optimization framework for benchmarking {MHD} flows over backward-facing steps},
  author={Spyridon Katsoudas and Grigorios Chrimatopoulos and Michalis Xenos and Efstratios Tzirtzilakis},
  journal={Mathematics},
  volume={13},
  number={18},
  pages={2953},
  year={2025},
  publisher={MDPI}
}
\end{document}